\documentclass{article}

\usepackage{microtype}
\usepackage{graphicx}
\usepackage{subfigure}
\usepackage{booktabs} 
\usepackage{placeins} 

\usepackage{hyperref}

 \usepackage[accepted]{icml2024}

\usepackage{amsmath}
\usepackage{amssymb}
\usepackage{mathtools}
\usepackage{amsthm}

\usepackage[capitalize,noabbrev]{cleveref}

\theoremstyle{plain}

\theoremstyle{definition}

\theoremstyle{remark}

\usepackage[textsize=tiny]{todonotes}

\icmltitlerunning{GPT-4 Generated Narratives of Life Events using a Structured Narrative Prompt: A Validation Study}

\begin{document}

\twocolumn[
\icmltitle{GPT-4 Generated Narratives of Life Events using a Structured \\
           Narrative Prompt: A Validation Study }



\icmlsetsymbol{equal}{*}

\begin{icmlauthorlist}
\icmlauthor{Christopher J. Lynch}{equal,yyy}
\icmlauthor{Erik Jensen}{equal,yyy,comp}
\icmlauthor{Madison H. Munro}{sch}
\icmlauthor{Virginia Zamponi}{yyy}
\icmlauthor{Joseph Martínez}{yyy,comp}
\icmlauthor{Kevin O'Brien}{yyy}
\icmlauthor{Brandon Feldhaus}{yyy}
\icmlauthor{Katherine Smith}{yyy}
\icmlauthor{Ann Marie Reinhold}{sch}
\icmlauthor{Ross Gore}{yyy}
\end{icmlauthorlist}

\icmlaffiliation{yyy}{Virginia Modeling, Analysis, and Simulation Center, Old Dominion University, Suffolk, Virginia, USA}
\icmlaffiliation{comp}{Electrical and Computer Engineering Department, Old Dominion University, Norfolk, Virginia, USA}
\icmlaffiliation{sch}{College of Engineering, Montana State University, Bozeman, Montana, USA}

\icmlcorrespondingauthor{Christopher J. Lynch}{cjlynch@odu.edu}

\icmlkeywords{Prompt Engineering, Large Language Models, ChatGPT, Machine Learning, Validation, Structure Narrative Prompt}

\vskip 0.3in
]



\printAffiliationsAndNotice{\icmlEqualContribution} 

\begin{abstract}
Large Language Models (LLMs) play a pivotal role in generating vast arrays of narratives, facilitating a systematic exploration of their effectiveness for communicating life events in narrative form. In this study, we employ a zero-shot structured narrative prompt to generate 24,000 narratives using OpenAI's GPT-4. From this dataset, we manually classify 2,880 narratives and evaluate their validity in conveying birth, death, hiring, and firing events. We observe that 87.43\% of the narratives sufficiently convey the intention of the structured narrative prompt. To automate the identification of valid and invalid narratives, we train and validate nine Machine Learning models on the classified datasets. Leveraging these models, we extend our analysis to predict the classifications of the remaining 21,120 narratives. The ML models all excelled at classifying valid narratives as valid, but experienced challenges at simultaneously classifying invalid narratives as invalid. Our findings advance the study of LLM capabilities, limitations, and validity and also offer practical insights for narrative generation and natural language processing applications.
\end{abstract}

\section{Introduction}
Large Language Models (LLMs) have emerged as powerful tools for generating text, crafting narratives, storytelling, and other forms of communication \cite{min2023recent, kalyan2023survey}. LLMs are capable of generating coherent and contextually relevant text across a variety of domains and scenarios. However, the quality and focused relevance of the produced narratives depend on the prompt provided to the language model. Prompt engineering plays a pivotal role in shaping LLM outputs and helping to generate messages that meet the intention of the prompt. Accountability, safety, honest use, and responsibility of LLM-generated results remain known challenges in using LLMs \cite{van2023chatgpt,sallam2023chatgpt,stokel-walker2023promise}, but prompt engineering may serve a useful avenue for addressing these challenges in narrative generation \cite{lynch2023structured}.

In this paper, we highlight the importance of prompt engineering and zero-shot learning in the context of narrative messaging with LLMs by exploring the role of machine learning (ML) models for automatically classifying narratives generated from an LLM ChatBot using a structured prompt and zero-shot learning. Prompt engineering enables researchers and practitioners to design prompts that guide LLMs to generate narratives aligned with specific themes, styles, or objectives. LLMs can be distracted by the inclusion of irrelevant context \cite{shi2023large}. By carefully crafting prompts, we can steer LLMs towards producing narratives that are engaging, coherent, and contextually relevant while yielding consistent and well-structured outcomes \cite{filippi2023measuring, lynch2023structured}. By leveraging auxiliary information provided in the prompts, zero-shot learning empowers LLMs to generate narratives for events or scenarios unseen during training, expanding their applicability to diverse storytelling tasks \cite{kojima2205large,wang2020generalizing}.

Narrative in science and health communication is effective and appealing for audiences across fields, topics, and mediums and helps to create openness to information \cite{dudley2023use}. Additionally, characters matter in evoking positive and negative audience experiences \cite{shanahan2019characters, barbour2016telling}. Narrative generation utilizing LLM ChatBots can work towards effectively merging character roles and science communications to produce more engaging language and connect emotionally and socially with the reader. Current work in this area includes (1) sentiment evaluation of ChatGPT 3.5-generated narrative messages compared to tweets which identified statistically indiscernible differences in sentiment levels in 4 of 44 evaluated sentiment traits \cite{lynch2023structured}, (2) that story weaving produced by an LLM only resulted in fewer logical flaws and was easier to understand than stories produced by an LLM in conjunction with humans \cite{zhao2023more}, and (3) LLM-generated messages for health awareness were statistically indistinguishable from human tweets with respect to sentiment, clarity, and semantic structure \cite{lim2023artificial}. Additionally, ChatGPT has been utilized to develop theme-relevant narratives for character driven simulation worlds \cite{johnson-bey2023toward} and for sifting story-worthy events from a collection of facts \cite{mendez2023using}.

Advancing this research area, we conduct a study to assess the validity of narratives generated by GPT-4 created through the use of a structured narrative prompt (SNP) via the methodology displayed in Figure 1. We generate 24,000 narrative messages using an existing SNP \cite{lynch2023structured} across birth, death, hiring, and firing events using a publicly available repository \cite{lynch2023mendeley}. We then sample 2,880 narratives and manually classify whether each narrative meets the intention of the prompt. Each narrative is independently assessed by two reviewers and then independently assessed by a third reviewer whenever a tie-breaker is needed. This classified dataset is utilized to train and test a series of ML models. For interpretability and transparency, we assess the validity of the ML models' precision on the classified data \cite{carvalho2019machine, lynch2021increased}. Next, we apply the trained ML models to predict the classifications of the remaining 21,120 data points and assess the agreement between the models' predictions. This research advances the study of LLM capabilities, limitations, and validity for creating natural language narratives using prompts and zero-shot learning while offering practical insights for narrative generation.

\begin{figure*}[ht]
\vskip 0.2in
\begin{center}
\centerline{\includegraphics[width=\columnwidth*2]{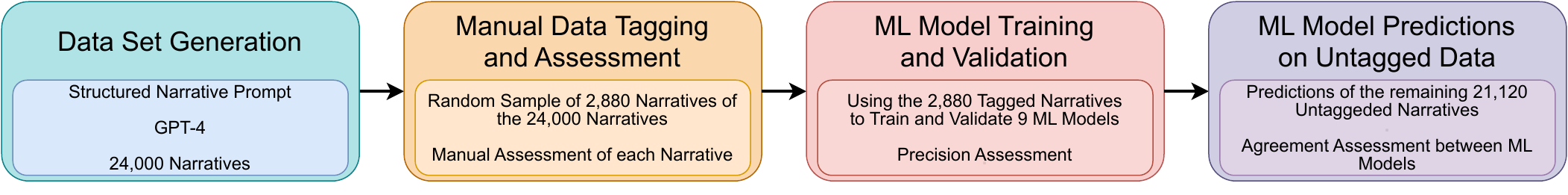}}
\caption{Research methodology. GPT-4 is prompted using an existing structured narrative prompt to produce 24,000 narratives across 4 life event types. These events undergo manual tagging to determine if each narrative meets the intention of its prompt. The tagged narratives are used to train and validate nine ML models. The validated ML models are then utilized to predict the classifications on the remaining 21,120 narratives.}
\label{icml-historical}
\end{center}
\vskip -0.2in
\end{figure*}

For baseline validity of the match between the generated narratives and the intention of their respective prompts, Table 1 provides the results of the manual classification of the ChatGPT-generated narrative messages. In total, the messages were found to sufficiently meets the specifications of the SNP in 87.43\% of cases. There exists a gap in the ability of ChatGPT to meet the intention of the prompt based on the type of life event, with only 72.08\% classified as \textit{Yes} for birth events but 96.67\% \textit{Yes} for hired events. However, this conveys strong evidence in support of using prompts to generate narratives from an LLM using zero-shot learning.

\begin{table}[t]
\caption{Classification results from manual data tagging for Birth, Death, Hired, and Fired events.}
\label{sample-table}
\vskip 0.15in
\begin{center}
\begin{small}
\begin{sc}
\begin{tabular}{lcccr}
\toprule
Event & Yes & No & n & Percent Yes \\
\midrule
Birth & 519 & 201 & 720 & 72.08\% \\
Death & 612 & 93  & 705 & 86.81\%  \\
Hired & 696 & 24  & 720 & 96.67\%  \\
Fired & 691 & 44  & 735 & 94.01\% \\
Total & 2518 & 362 & 2880 & 87.43\% (95\% CI $\pm$ 0.40) \\
\bottomrule
\end{tabular}
\end{sc}
\end{small}
\end{center}
\vskip -0.1in
\end{table}

\subsection*{\textbf{Key Takeaways:}}
\begin{itemize}
\item By leveraging prompt engineering and auxiliary information provided in prompts, LLMs can effectively generate contextually relevant narratives across a diverse range of topics.
\begin{itemize}
    \item ChatGPT GPT-4 averaged 87.43\% valid narrative based on the SNP inputs.
\end{itemize}
\item Precision results of the trained ML models indicates usefulness of various ML techniques and highlights timing concerns with message prediction pertaining to scalability of data for some techniques. 
\item Agreement between ML models trained on generated narrative data can be utilized to automatically classify future messages.
\begin{itemize}
    \item The LLM ChatBot can utilize an ensemble of trained ML models to assess and refine future narrative generation.
    \item Erroneous narratives can be automatically filtered.
\end{itemize}
\end{itemize}

\section{Related Work}
\textbf{Zero-shot learning} has gained significant attention in the field of ML, particularly for its applications in natural language processing and text generation tasks. Models are trained to generalize to classes that are unseen during training, instead, leveraging auxiliary information that is provided at inference time. Thus, enabling language models to adapt to novel scenarios or domains without requiring explicit examples for every task. In the context of LLMs, zero-shot learning techniques have been employed to extend the capabilities of text generation models by allowing them to generate narratives for events or scenarios not encountered during training \cite{lim2023artificial, lynch2023structured, mesko2023prompt} and for generating sequences of actionable tasks \cite{huang2022language}. Instruction tuning has also been shown to be successful at improving zero-shot performance \cite{wei2021finetuned} as well as using unlabeled data to co-train a prompted model has also been shown to provide performance improvements under the right conditions \cite{lang2022co}. Additionally, zero-shot learning has been explored to expand into the use of chain-of-thought prompting \cite{kojima2205large}. 

\textbf{Prompt engineering} has emerged as a methodology for guiding LLMs in generating coherent and contextually relevant text using explicit prompt structures that provide elements such as instructions, context, input data, and output indicators \cite{giray2023prompt}. Prompts serve as templates to influence the outputs from pre-trained language models using textual strings that allow the language models to solve numerous tasks \cite{brown2020language, liu2023pre, kalyan2023survey}. This allows LLMs to generate text that aligns with specific themes, styles, and objectives. Prompt engineering techniques range from simple prompts providing high-level guidance to complex prompts that incorporate constraints.

The Goal Prompt Evaluation Iteration (GPEI) methodology incorporates data inclusion and principles from explainable AI to promote transparency and justifiability in LLM responses \cite{velasquez2023prompt}. Increasing prompt specificity can lead to intensified neutrality in ChatGPT responses \cite{henrickson2023prompting}. Recommendations for improving prompt usefulness have also included asking ChatGPT to enter the process and make recommendations for its own prompts \cite{mesko2023prompt}.

\textbf{Narrative generation tasks} utilize prompt engineering as a pivotal component in shaping narrative's content, structure, and coherence. Researchers have explored various approaches to prompt engineering, including the use of SNPs \cite{lynch2023structured}, conditional generation techniques \cite{liu2023pre}, and prompt fine-tuning strategies \cite{wei2021finetuned}. These approaches enable LLMs to produce narratives that meet the intended criteria while exhibiting desirable properties, such as consistency, realism, and contextual relevance.

\section{Experimentation}
Experimentation is carried out using the four steps presented in Figure 1. The process starts with generating the data, then manual assessment of a sample of the generated narratives, then ML model training and validation of a suite of ML models, and ending with predicting the classifications of the remaining untagged narratives and an assessment across the validated ML models. Our experimental system consists of an AMD Ryzen 5 3600, 6-core machine, with 8 GB DDR4 RAM, and an NVIDIA GeForce RTX 3070 GPU, running Linux Manjaro. All BERT and Keras runs utilize the GPU.

\subsection{Preparing the Data}
An existing SNP \cite{lynch2023structured} was utilized to prompt OpenAI's GPT-4 model using OpenAI's ChatGPT API \cite{openAI2023chatGPT}. The SNP was utilized to create 24,000 narratives based on 4 life event types, birth, death, hiring, and firing. The SNP was populated with individualized data associated with these four event types from simulated agents reusing code from a publicly available repository \cite{lynch2023mendeley}. A random sample of 12\% of the narratives were then pulled to form a classification set. This resulted in 2,880 narratives to be tagged. The \textit{n} column of Table 1 provides the resulting number of samples with respect to each event type.

\begin{figure}[ht]
\vskip 0.2in
\begin{center}
\centerline{\includegraphics[width=\columnwidth]{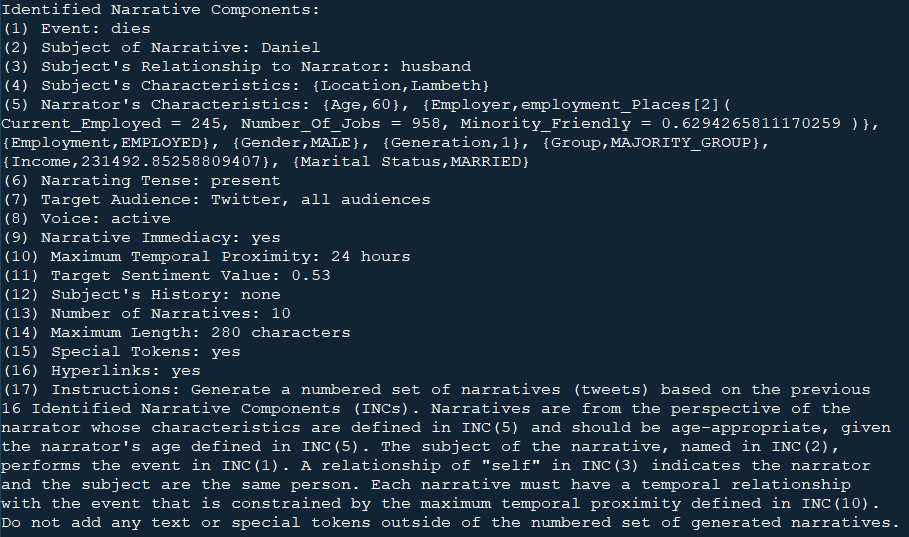}}
\caption{Sample Structured Narrative Prompt setup utilized to prompt the LLM. Each prompt sent to the LLM contains unique information in fields 2-5 as they pertain to the subject and narrator characteristics.}
\label{icml-historical}
\end{center}
\vskip -0.2in
\end{figure}

\subsection{Manual Data Tagging}
The 2,880 narratives selected for classification were randomly divided amongst two groups of four reviewers each. Each narrative was assigned two reviewers, one from each group. One remaining person was assign as a tie-breaker for any tied decision, for a total of nine reviewers. Reviewers were provided the specific prompt utilized to create the narrative as well as the response produced by the LLM. Reviewers were then asked to provide a binary assessment \textit{Yes} or \textit{No} tag for each narrative. A tag of \textit{Yes} if they felt that the narrative met the intention of the prompt or a tag of \textit{No} otherwise. A predefined list of exclusionary criteria were provided to the reviewers, including: (1) wrong event; (2) subject (target person) of narrative is wrong; (3) wrong subject-narrator relationship; (4) incorrect narrator or subject characteristics; (5) temporal error; or (6) narrative is not age-appropriate given age of narrator. Reviewers also selected 1 or more exclusionary criteria if tagging a narrative with \textit{No}; however, an exploration of exclusionary rational was not conducted as part of this study.

All reviews, including tie-breakers, were conducted over a two-week period. An automated form was provided to the reviewers to help expedite the review process. For each narrative provided to the reviewer, this form presented the event type, the instruction for the reviewers (the same for every narrative), the information on the characteristics of the simulated agent provided in the SNP to the LLM (this is the only variable information that differs across the input prompts), and the resulting narrative provided by ChatGPT. Figure 2 provides a sample representation of an untagged narrative during the review process.

\begin{figure}[ht]
\vskip 0.2in
\begin{center}
\centerline{\includegraphics[width=\columnwidth]{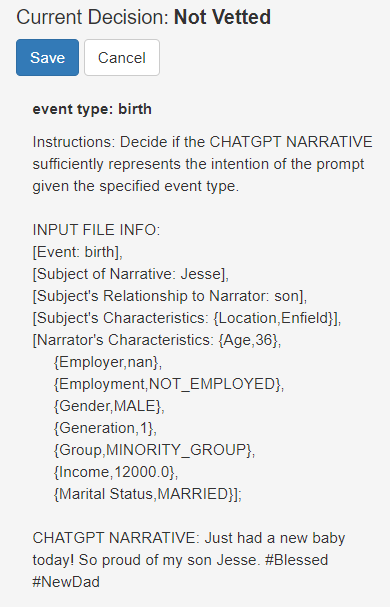}}
\caption{Sample user interface during manual data tagging. Narratives start untagged and reviewers independently provide binary Yes/No tags for each narrative in their respective sets. Ties among reviewers are broken by an independent third party.}
\label{icml-historical}
\end{center}
\vskip -0.2in
\end{figure}

All reviews were conducted independently and a review coordinator handled the automated aggregation of responses.  During this process, reviewer names were made anonymous. Any narrative receiving both a \textit{Yes} and a \textit{No} vote were exported and sent to the tie-breaker for a final decision. The tie-broken set was then folded back into the result set. Every narrative receiving two \textit{Yes} votes received a classification of Yes and every narrative receiving two \textit{No} votes received a final classification of \textit{No}. In total, only 295 narratives (10.24\%) required tie-breaking. Table 1 provides the results of the aggregated data tagging for all narratives as well as per event category.

\subsection{Model Generation using Tagged Data}
Nine ML models, each selected for its unique capabilities and architectures, were selected to train and validate on the tagged data set, including Random Forest \cite{biau2016random}, support vector machine (SVM) \cite{chauhan2019problem}, eXtreme Gradient Boosting \cite{zhang2023insights}, and various Keras layers such as Long Short-Term Memory (LSTM) \cite{yu2019review}, Gated Recurrent Unit (GRU) \cite{irie2016lstm}, Rectified Linear Unit (RELU) \cite{rasamoelina2020review}. Additionally, Bidirectional Encoder Representations from Transformers (BERT) \cite{jin2020bert, zhang2020semantics, acheampong2021transformer} were employed with multiple configurations of token limits (64, 128, and 256) for input sequences. To ensure robustness, each model underwent 10 K-fold cross-validation, splitting the data into 10 groups for iterative training and validation. 

A 2x2 confusion matrix is constructed for each of the ML models utilizing the average of its k-fold tested models. Fisher's exact test is applied to the confusion matrix to identify statistically significant differences between each model's binary classifications of the tagged data and the actual tagged classifications. The Fisher's exact test is appropriate when very small sample sizes exist within any of the cells of the contingency table \cite{upton1992fisher,bower2003use}. The models are also tested using the McNemar test to determine if statistically significant differences exist in the distributions of the Yes/No variables \cite{pembury2020effective} and provide further insight into the models' performance and capabilities. The null hypothesis for the Fisher's exact test is that a significant difference exists in the distribution of \textit{Yes}/\textit{No} classifications between the model's binary classifications and the actual classifications. The alternative hypothesis is that no difference exists between the distribution of \textit{Yes}/\textit{No} classifications made by the model and the actual classifications.

\subsection{Model Prediction on Remaining Untagged Data}
Each of the nine models constructed using the 2,880 tagged data points is utilized to predict the classification of the 21,120 data points comprising the originally untagged data points. This results in nine sets of predictions on the untagged data. McNemar tests are again used to determine if  statistically significant differences exist in the distributions of the \textit{Yes}/\textit{No} variables for the predicted classifications between the models. However, in this case the true classifications are not know. Therefore, to form a comparison point for assessing the results of the predictions, we construct an ensemble hypothesis consisting of binary \textit{Yes}/\textit{No} classifications for each of the 21,120 untagged narratives. Each narrative classifies as \textit{Yes} that at least 5 of the models predicted that the narrative classifies as \textit{Yes}; otherwise, the narrative classifies as \textit{No}. An informal sanity check on the ensemble results was conducted via visual inspection of a small sample of both the \textit{Yes} and \textit{No} classified narratives. The checked narratives within this visually inspection matched the classification that the human inspector would have assigned.

\section{Results}
The results and primary findings related to assessing the validity of the SNP, the significance of the models, and the significance of the predicted classifications of the untagged data are presented in this section.

\subsection{Structured Narrative Prompt Validity}
The success of the utilized SNP is evident through the achieved aggregated accuracy of 87.43\% (Table 1) across all narratives as evaluated through manual tagging. This high level of accuracy indicates that the narratives generated by GPT-4 in response to the structured prompts effectively conveyed the intention of the prompt throughout a majority of the narratives across various life events, including birth, death, hiring, and firing. The structured prompt provided clear guidance and constraints, enabling the model to produce narratives that aligned closely with the specified themes and objectives. 

This result underscores the effectiveness of both prompt engineering and zero-shot learning in shaping the output of LLMs while also maintaining coherence and contextual relevance. The high accuracy also highlights the potential of structured narrative prompts to enhance the quality and consistency of narrative generation and to facilitate meaningful communication and foster engagement across domains. However, the range in the level of accuracy across event types [72.08, 96.67] indicates that not all life event narrations can be assumed to be equally effectively generated by LLMs.

\subsection{ML Model Validity}
Each of the nine models undergo assessment through two primary avenues: (1) Fisher's exact test, and (2) positive/negative classification precision. As outlined in the methodology, Fisher's exact test is applied to compare each model to its baseline classifications. In this case, the manually tagged data set. The models' binary classifications are evaluated against the manually classified data set to determine their accuracy at meeting the intention of the SNP. The statistical analysis helps to identify inconsistencies in the models' predictive capabilities. By subjecting the models to both binary precision testing and Fisher's exact test, we attain a robust evaluation process that enhances the confidence and validity of the findings. 

Figure 4 provides the results of the Fisher's exact tests applied to all nine models across the four life event types. Using a significance level of 0.05 (represented as the red dashed line in the plots), 29 of the 36 tests achieved statistically significant results as a result of their training and validation on the tagged data set. This indicates that there is evidence in support of rejecting the null hypothesis in favor of the alternative hypothesis that there is not a significant difference between the models' binary classifications and the actual classifications. Note, in all 7 of the cases where the P-values were greater than 0.05, the P-values are equal to 1. This is an artifact of making zero \textit{No} classifications and does not indicate failing to reject the null hypothesis.

\begin{figure*}[ht]
\vskip 0.2in
\centerline{
  \subfigure(a){\includegraphics[width=0.45\textwidth, height=150pt]{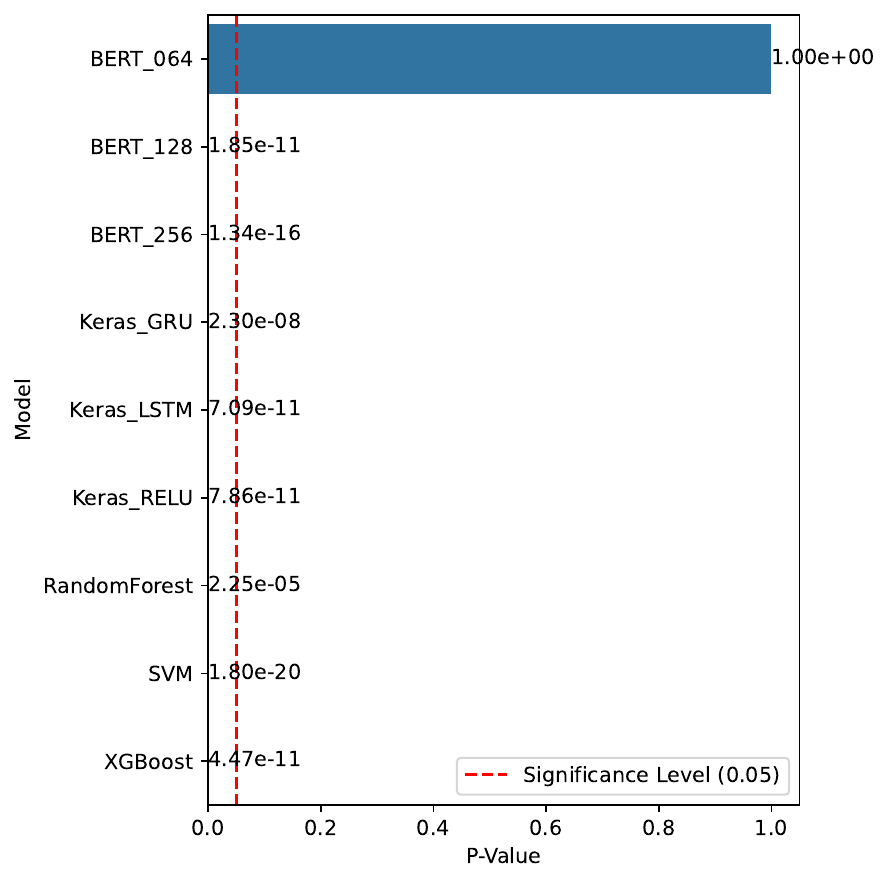}} 
  \subfigure(b){\includegraphics[width=0.45\textwidth, height=150pt]{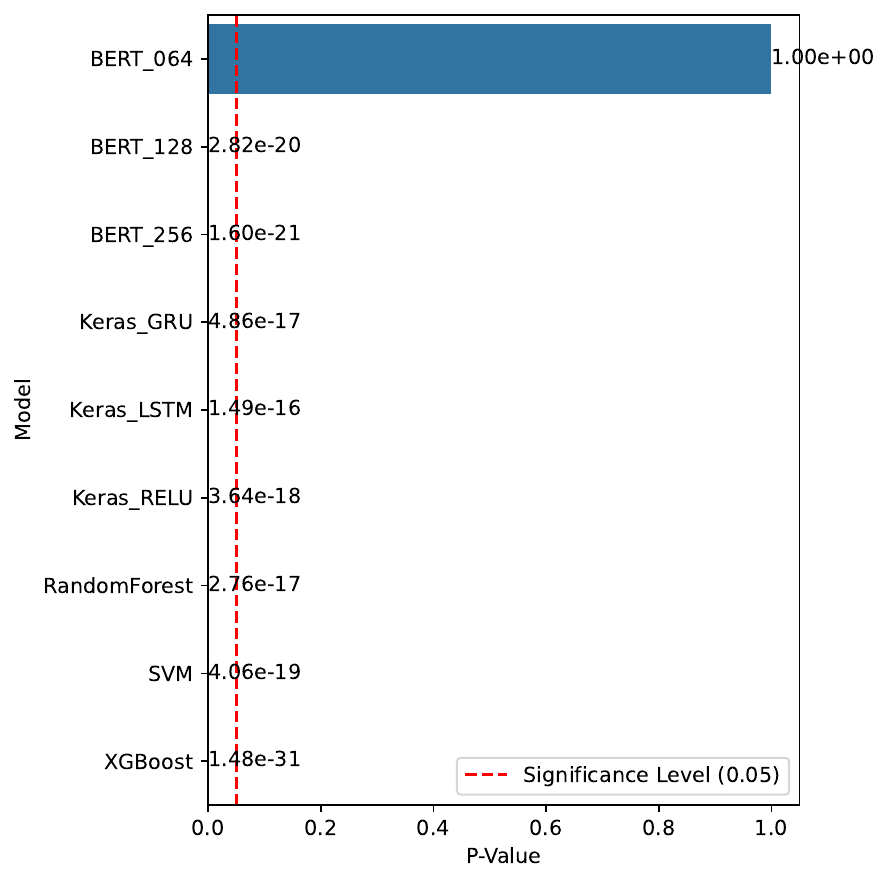}} 
  }
\centerline{
\subfigure(c){\includegraphics[width=0.45\textwidth, height=150pt]{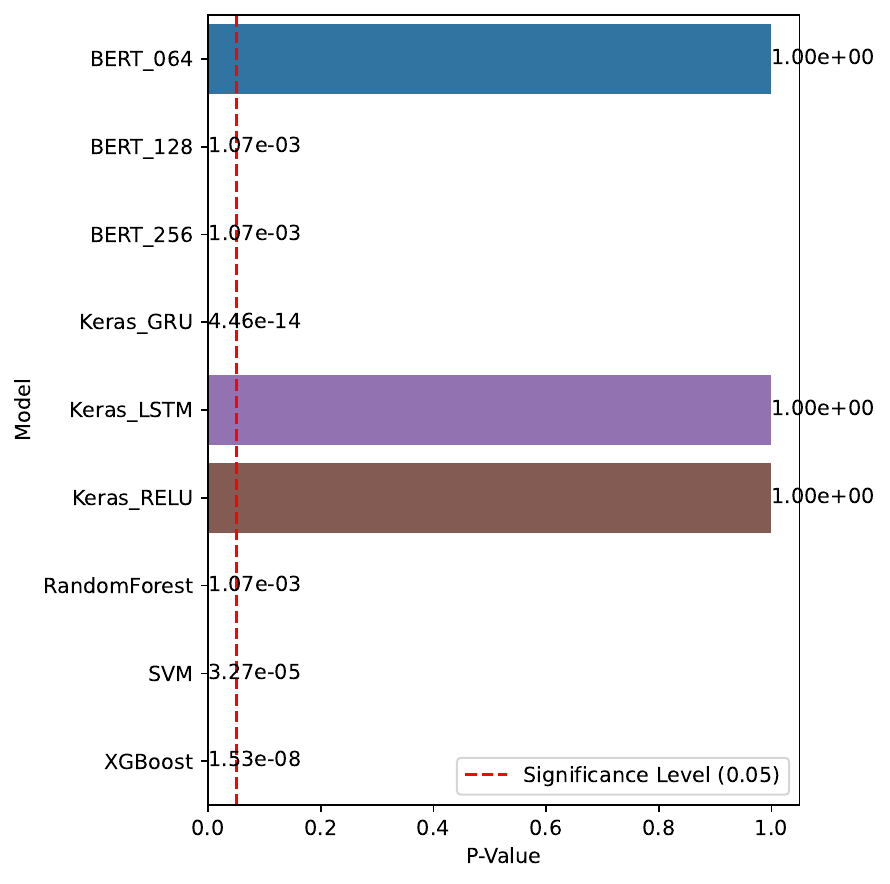}}
  \subfigure(d){\includegraphics[width=0.45\textwidth, height=150pt]{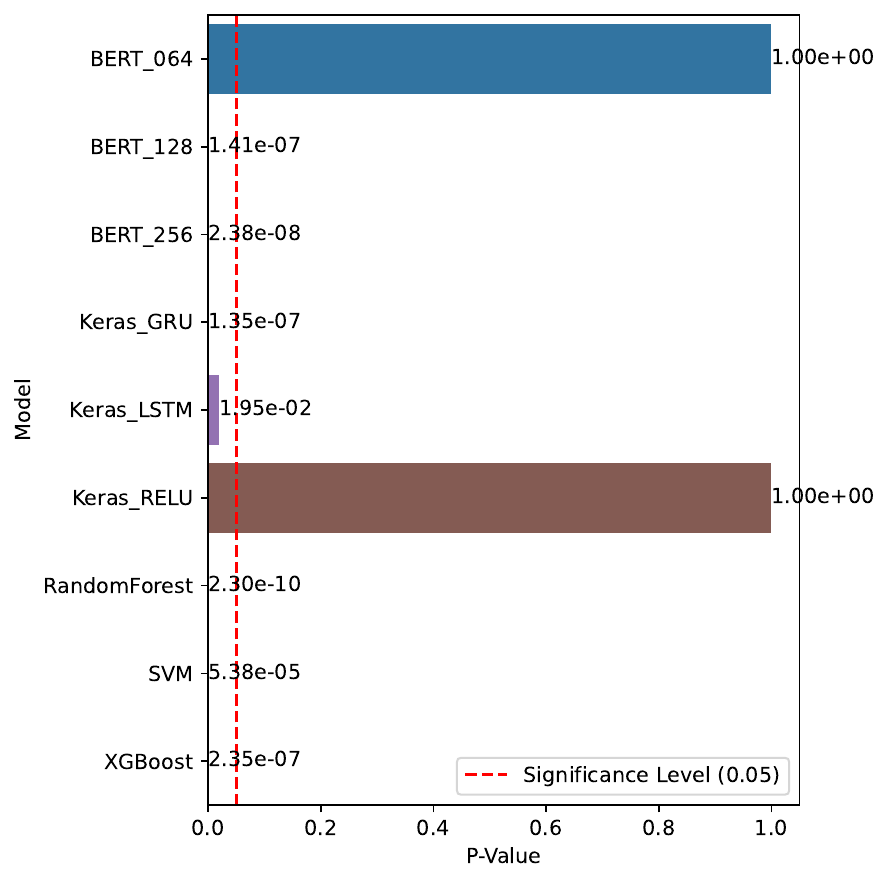}}
}
  \caption{Fisher's exact test results on the confusion matrices for each model. P-values of 1.0 occur in cases where 0 \textit{No} classifications occurred. Results are grouped by event type, (a) Birth event narratives, (b) Death event narratives, (c) Hired event narratives, and (d) Fired event narratives.}
\label{fig:foobar}
\vskip -0.2in
\end{figure*}

These statistically significant results support that the SNP can successfully and consistently yield narratives matching the intention of the prompts. Next, the precision of the models is assessed with respect to how well the models were able to classify the tagged narratives given that the answers to the classifications are known. Figure 5 provides the positive (\textit{Yes}) and negative (\textit{No}) precision values for each model across each life event type.

\begin{figure*}[ht]
\vskip 0.2in
\centerline{
  \subfigure(a){\includegraphics[width=0.5\textwidth, height=150pt]{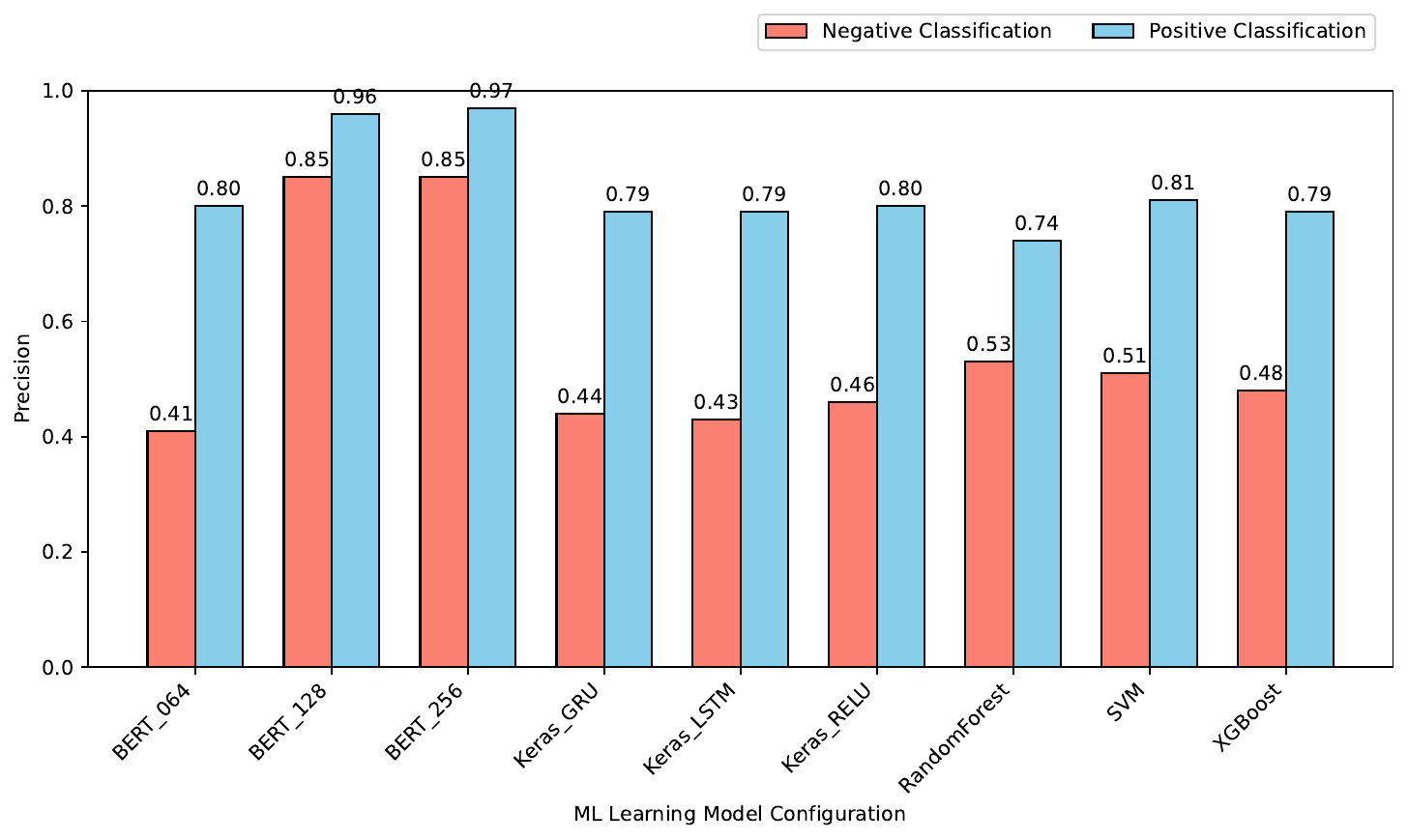}} 
  \subfigure(b){\includegraphics[width=0.5\textwidth, height=150pt]{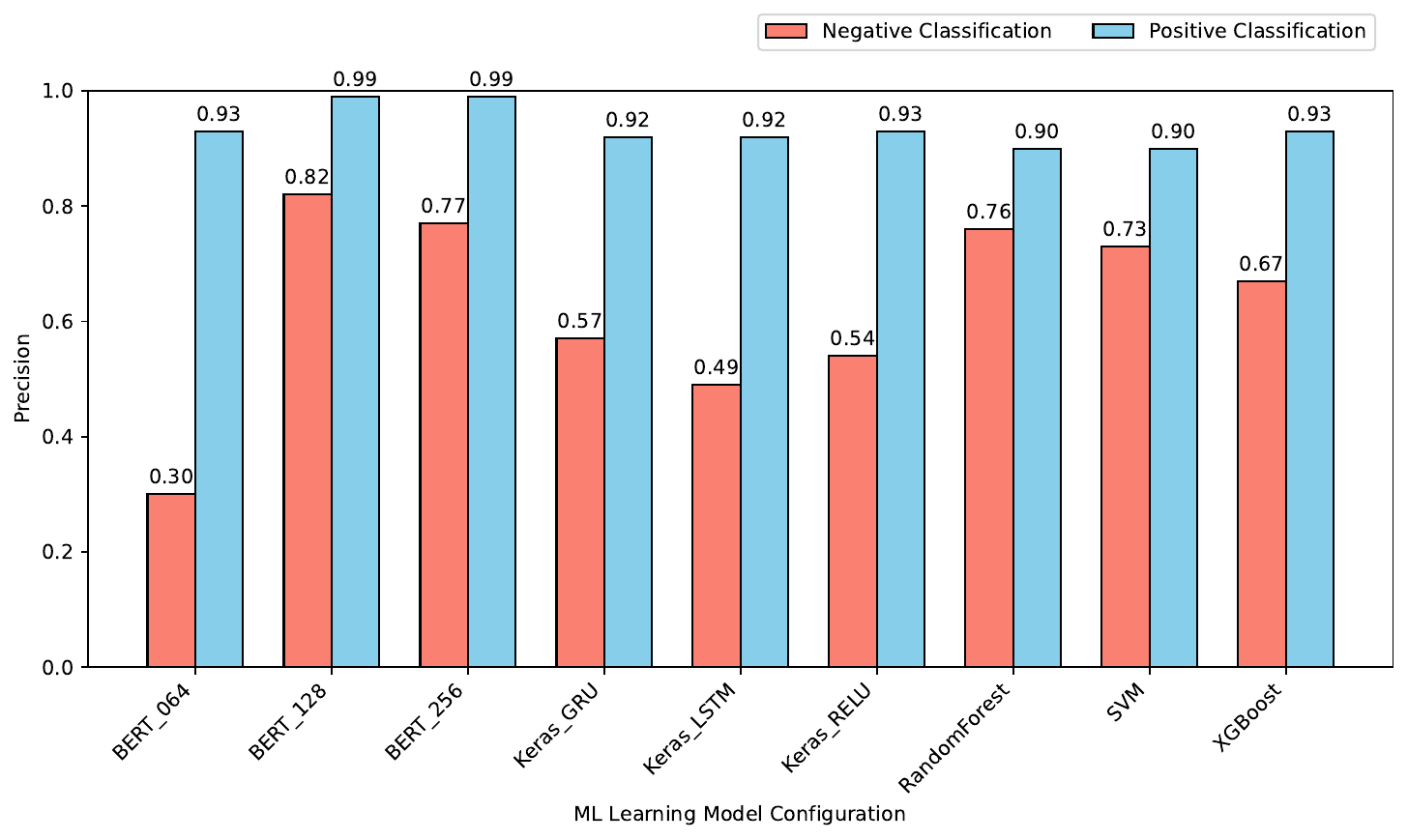}} 
  }
\centerline{
\subfigure(c){\includegraphics[width=0.5\textwidth, height=150pt]{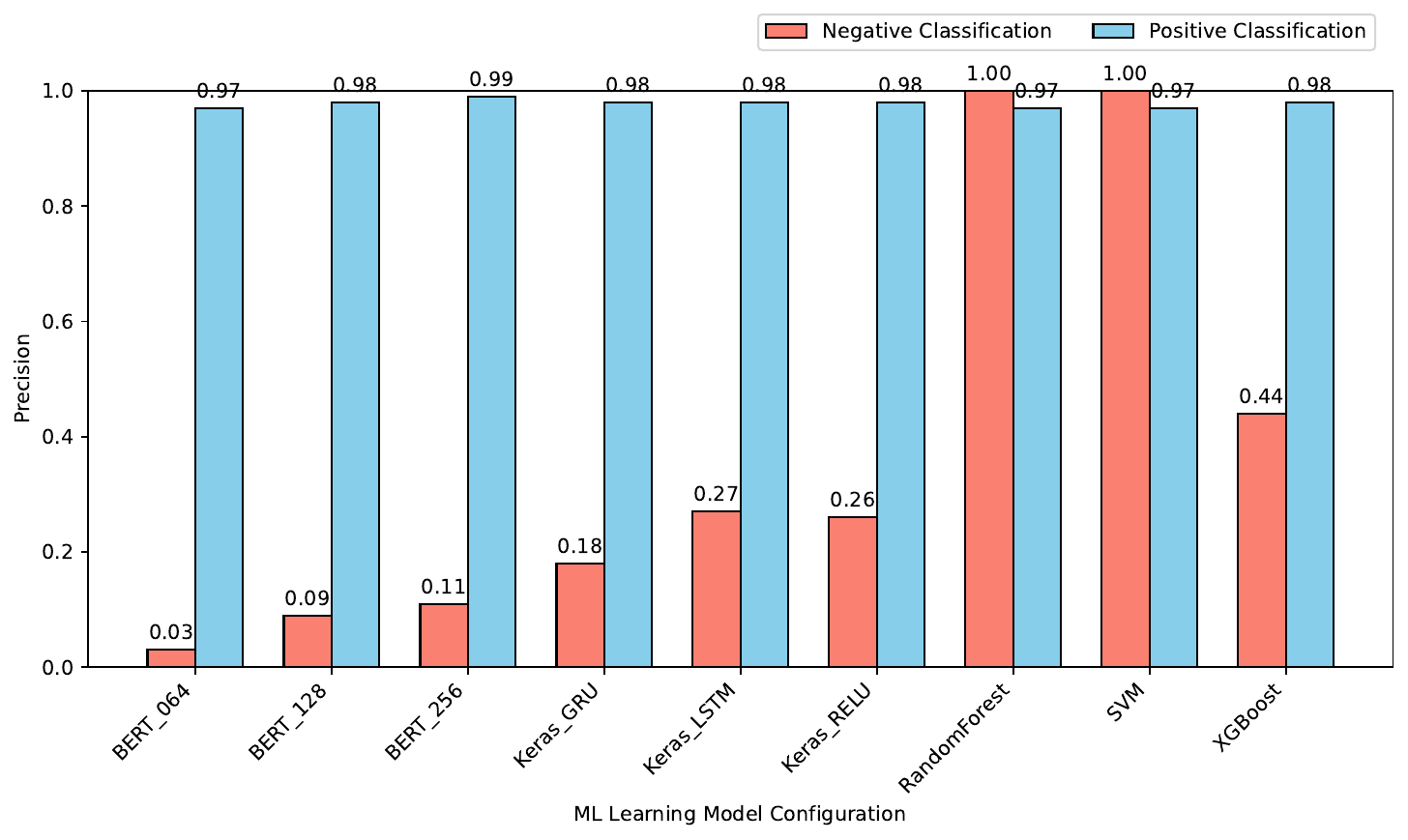}}
  \subfigure(d){\includegraphics[width=0.5\textwidth, height=150pt]{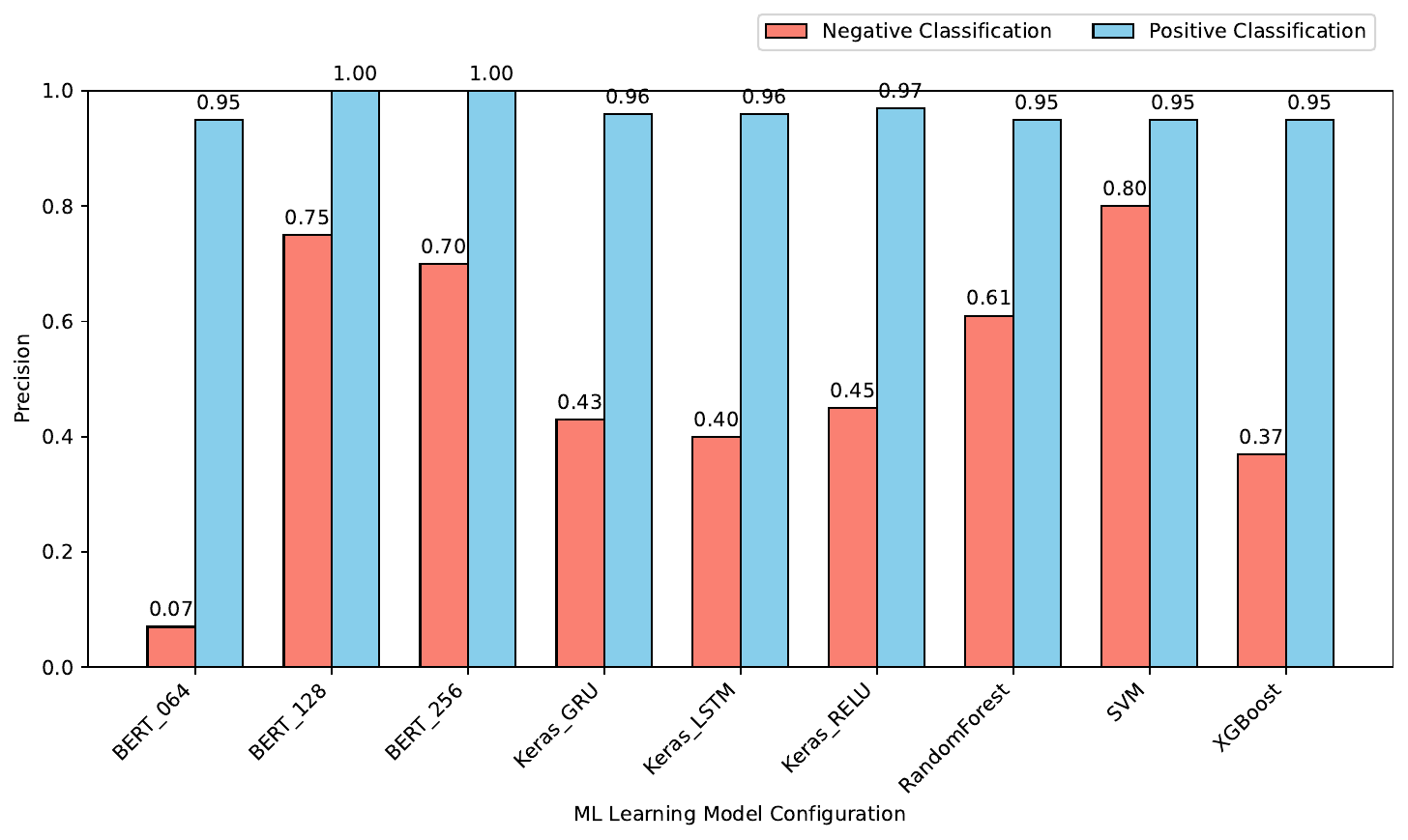}}
}
  \caption{ML models' binary \textit{Yes} (blue) / \textit{No} (red) classification precision for (a) Birth event narratives, (b) Death event narratives, (c) Hired event narratives, and (d) Fired event narratives.}
\label{fig:foobar}
\vskip -0.2in
\end{figure*}

Almost every model performs better at \textit{Yes} classifications that \textit{No} classifications across all four life event types. Generally, each model displays a sizable difference between their \textit{Yes} and \textit{No} precisions. A notable exception exists for the Random Forest and SVM models for hired events, where the precision exceeded 95\% for positive and negative classifications. A challenge experienced throughout this process was the low sample size of \textit{No} classifications within the tagged data set. This made it more difficult during training for a model to determine how to accurately classify \textit{No} messages. This is discussed further within the Limitations section.

\subsection{ML Model Prediction Validity}
An additional validation avenue is explored utilizing agreement matrices that compare the distribution of binary classifications made by each model on the untagged data set (n = 21,120) compared to each other models' classification predictions for each of the life event types. For these tests, each of the nine models is trained using all of the tagged data (n = 2,880), still separated by life event type. These models are then used to predict the classifications on the untagged data. The agreement matrices measure agreement through the proportion of matches (\textit{Yes} to \textit{Yes} and \textit{No} to \textit{No}) between models. 

The null hypothesis  is that there does not exist a statistically significant difference in distributions of the binary classifications made by each model. P-values less than 0.05 indicate that evidence exists to support rejecting the null hypothesis in favor the alternative hypothesis that a difference between the distribution of binary classifications does exist. Figure 7, parts a-d (Appendix A), provides the agreement matrix results for each life event type. In general, statistically significant outcomes are observed in almost all instances in the matrices for the birth, death, and hired life event narratives. However, far fewer statistically significant outcomes are observed with respect to the fired narratives.

Next, McNemar tests are conducted with the models' predicted classifications of the untagged data (n = 21,120) using an ensemble hypothesis comparison set where baseline \textit{Yes} classifications are the result of 5+ models predicting a \textit{Yes}. The null hypothesis tested in these cases states that there is no statistically significant difference in distributions of the binary classifications made by each model. P-values less than 0.05 indicate that evidence exists to support rejecting the null hypothesis in favor the alternative hypothesis that a difference between the distribution of binary classifications does exist. Appendix C provides the figures showcasing the results of the McNemar tests for significant differences utilizing the ensemble hypothesis comparison. A majority of the assessments are statistically significant with P-values less than the 0.05 significance level. These instances provide evidence supporting the rejection of the null hypothesis in favor of the alternative hypothesis. In these cases, a statistically significant difference in the distribution of \textit{Yes}/\textit{No} classifications between model pairs exists between the compared ML models.

\subsection{Timing Considerations for ML Building and Predicting}
To conclude our exploration, we assessed the computational efficiency of our ML models. This included measuring both the training time, in seconds, required to train the ML models, as well as the inference time, also in seconds, needed to apply the trained models for predicting the classifications of the untagged data. Figure 6 provides the times associated with both of these efficiency measures averaged across all four life event types.

\begin{figure}[ht]
\vskip 0.2in
\begin{center}
\centerline{\includegraphics[width=\columnwidth]{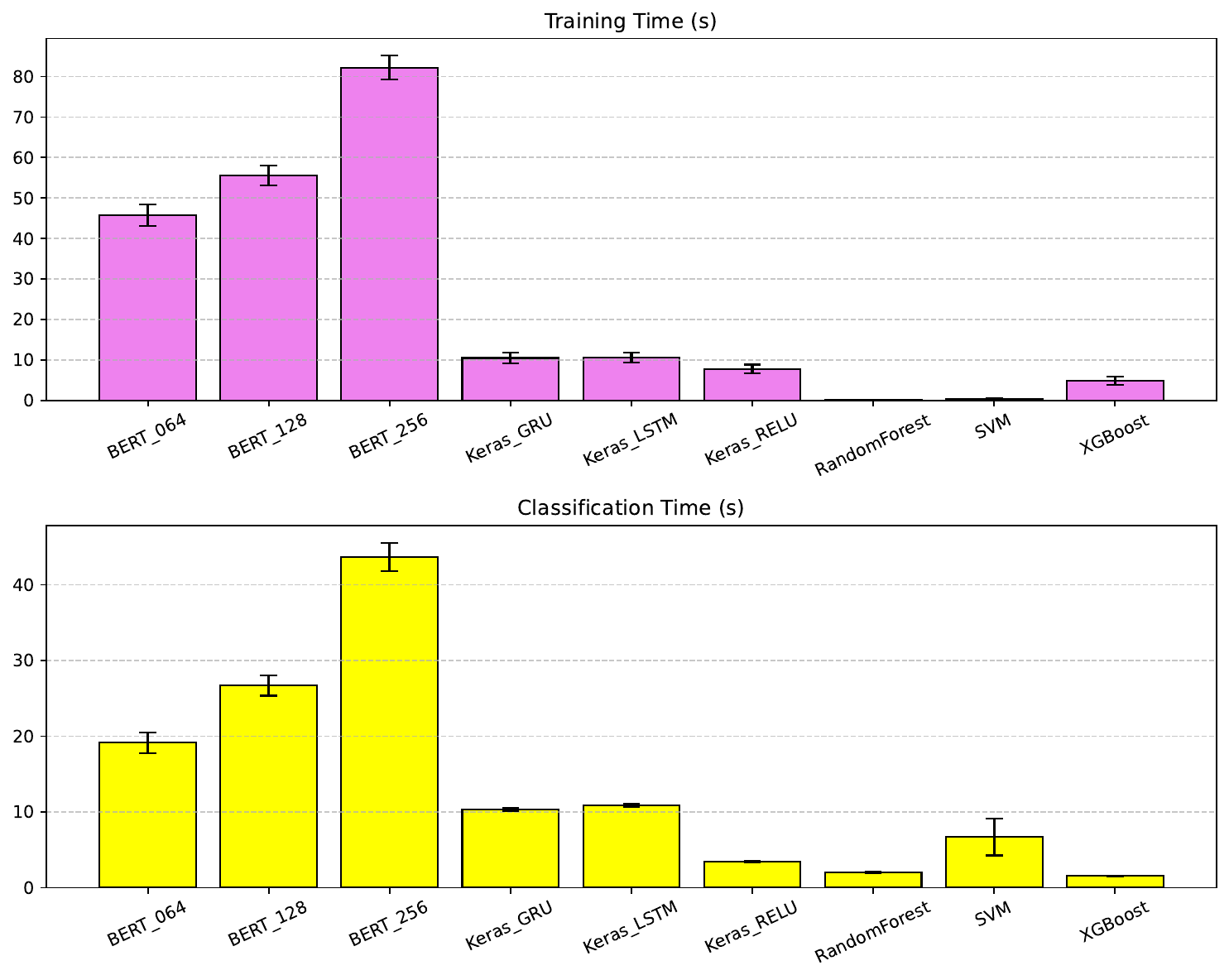}}
\caption{Average training and classification times in seconds, per model. Training is conducted from the tagged data set (n = 2,880) and classification time is measured over the untagged data set (n = 21,120). All event types are averaged together.}
\label{icml-historical}
\end{center}
\vskip -0.2in
\end{figure}

The timing values paired with the precision figures support that the traditional ML models of Random Forests and SVMs are fitting starting points in the exploration process. Both of these models trained and ran predictions very quickly while also achieving very good results in assigning \textit{Yes} and \textit{No} classifications. The Keras and BERT model configurations were great fits for the binary classifications but, particularly for BERT, display alarming timing trends if scaling the size of the tagged data sets used for training.

\section{Study Limitations}
While the study sheds light on narrative generation across various life event types and enhances transparency in prompting narratives from LLMs, several limitations warrant consideration. Firstly, the generalizability of the results may be constrained by the specific life event types evaluated in this study, namely birth, death, hired, and fired events. The effectiveness of SNPs and ML models in generating narratives for other life event categories remains uncertain and necessitates further investigation. Additionally, the reliance on manual tagging for evaluation introduces the potential for subjective biases and inconsistencies, which may affect result accuracy. However, the inclusion of multiple reviewers for each narrative aims to mitigate this issue. Furthermore, the choice of ML models and parameters may influence performance and generalizability. Future research endeavors should aim to explore a broader range of life event types and employ standardized evaluation methodologies to bolster the reliability and applicability of the findings across diverse domains and scenarios.

Additionally, our model training and validation efforts were impacted by the class imbalance problem \cite{ali2013classification,megahed2021class} as a result of the low number of manually classified \textit{No} cases within the tagged narrative set. While this was a great representation of GPT-4 to properly respond to the SNP, this presented early challenges in training the ML models as many of the models pushed towards an always classify as \textit{Yes} solution. Creating a larger data set of manually tagged data points could have helped in increasing the number of \textit{No} tags; however, additional time and resources were not available to allow for additional tagging. This prompted the expansion of models beyond the initial set of purely ML models to include the three Keras and three BERT models. This allowed us to conduct a more robust exploration of the problem space. Furthermore, the timing testing indicates that larger sets of tagged data may warrant further testing and exploration within the realm of SNPs.

\section{Conclusion}
The statistically significant results from our analysis serve as compelling evidence affirming the effectiveness of the utilized SNP in guiding narrative generation by LLMs. With 29 out of 36 tests yielding significant outcomes at a 0.05 significance level for the ML models' validity, our study underscores the SNP's prowess in consistently eliciting narratives that closely align with the intention of the prompts. These findings not only validate the reliability of our approach but also shed light on the remarkable capacity of LLMs to comprehend and adhere to structured guidelines when crafting narratives. The pipeline of utilizing a SNP to yield narratives, manually classifying the narratives, and applying ML modeling to the classification of the narratives appears a fruitful path towards automating the classification process and allowing for the future potential of a recursive loop where the automatic evaluation of generated narratives can also be utilized to improve upon the prompt and enhance the narrative generation process.

By demonstrating the SNP's ability to yield narratives that resonate with human intent, our research opens doors to a myriad of applications, from enhancing storytelling in artificial intelligence to generating effective and empathetic narratives from simulation agents to fueling impassioned communication between policy makers and their constituents. Ultimately, our study highlights the transformative potential of structured narrative prompts in transparently harnessing the power of LLMs to communicate with depth, clarity, and authenticity.

\section*{Software and Data}
The data and code utilized by this research study have been uploaded as "Supplementary Material". This data will be moved to a publicly accessible repository if accepted for publication.

\section*{Broader Impact}
This paper presents work whose goal is to advance the field of Machine Learning. There are many potential societal consequences of our work, none which we feel must be specifically highlighted here.

\section*{Acknowledgements}
This study was funded, in part, by the Office of Enterprise Research and Innovation at Old Dominion University (\#300916-010). This work was also supported, in part, by the Commonwealth Cyber Initiative (CCI), an investment in the advancement of cyber R\&D, innovation, and workforce development. For more information about
CCI, visit www.cyberinitiative.org.

\bibliography{references}
\bibliographystyle{icml2024}

\newpage
\appendix
\onecolumn
\section{Agreement matrices of ML models' predicted classifications.}
The agreement matrices provided in Figure 7 assesses the consistency between the binary classification of the ML models. Each row and column reflects one of the ML models and each cell represents the level of agreement between the two corresponding models. A significance level of 0.05 is utilized to reflect statistically significant outcomes within the tables.

\begin{figure*}[ht]
\vskip 0.2in
\centerline{
  \subfigure(a){\includegraphics[width=0.5\textwidth, height=150pt]{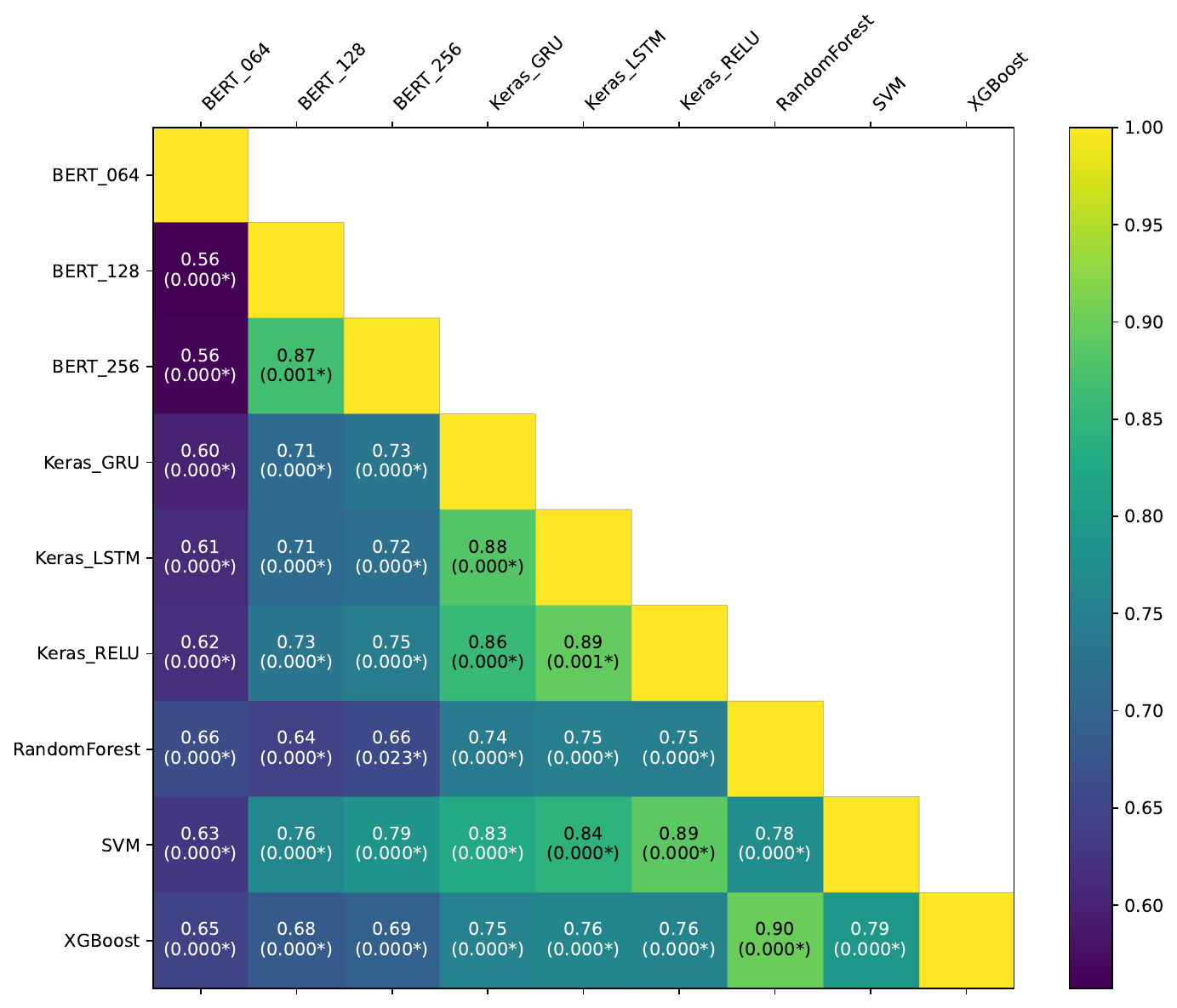}} 
  \subfigure(b){\includegraphics[width=0.5\textwidth, height=150pt]{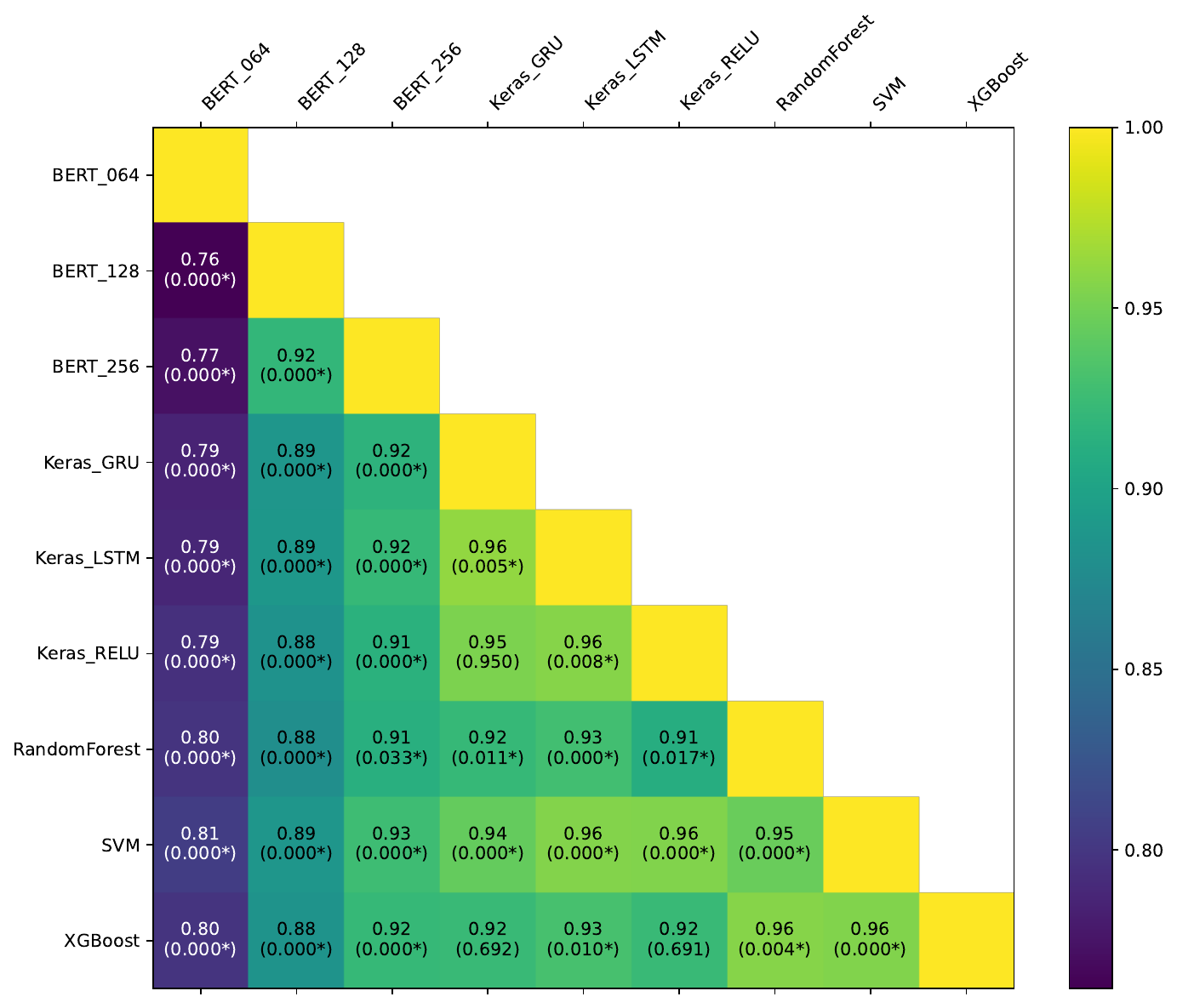}} 
  }
\centerline{
\subfigure(c){\includegraphics[width=0.5\textwidth, height=150pt]{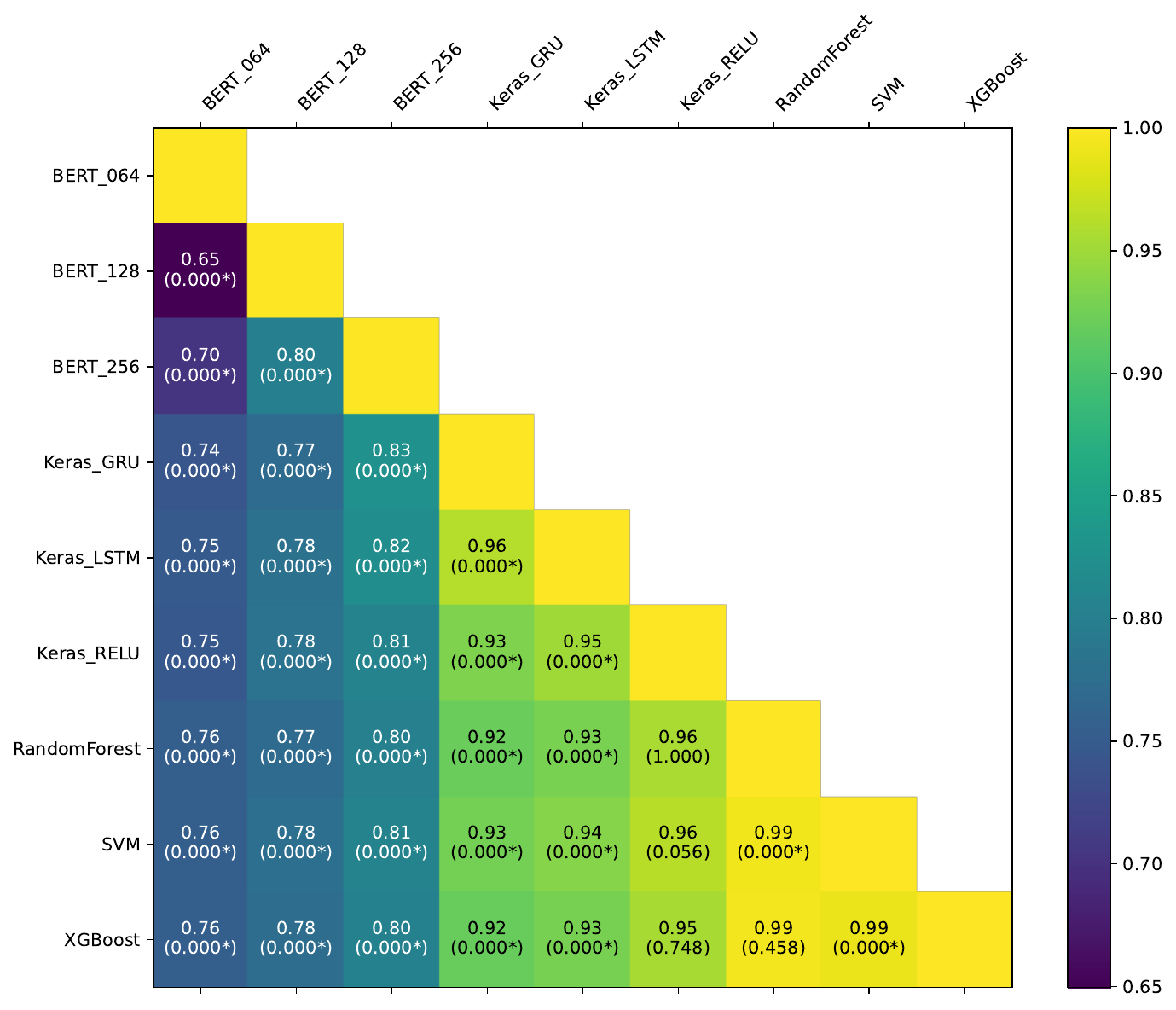}}
  \subfigure(d){\includegraphics[width=0.5\textwidth, height=150pt]{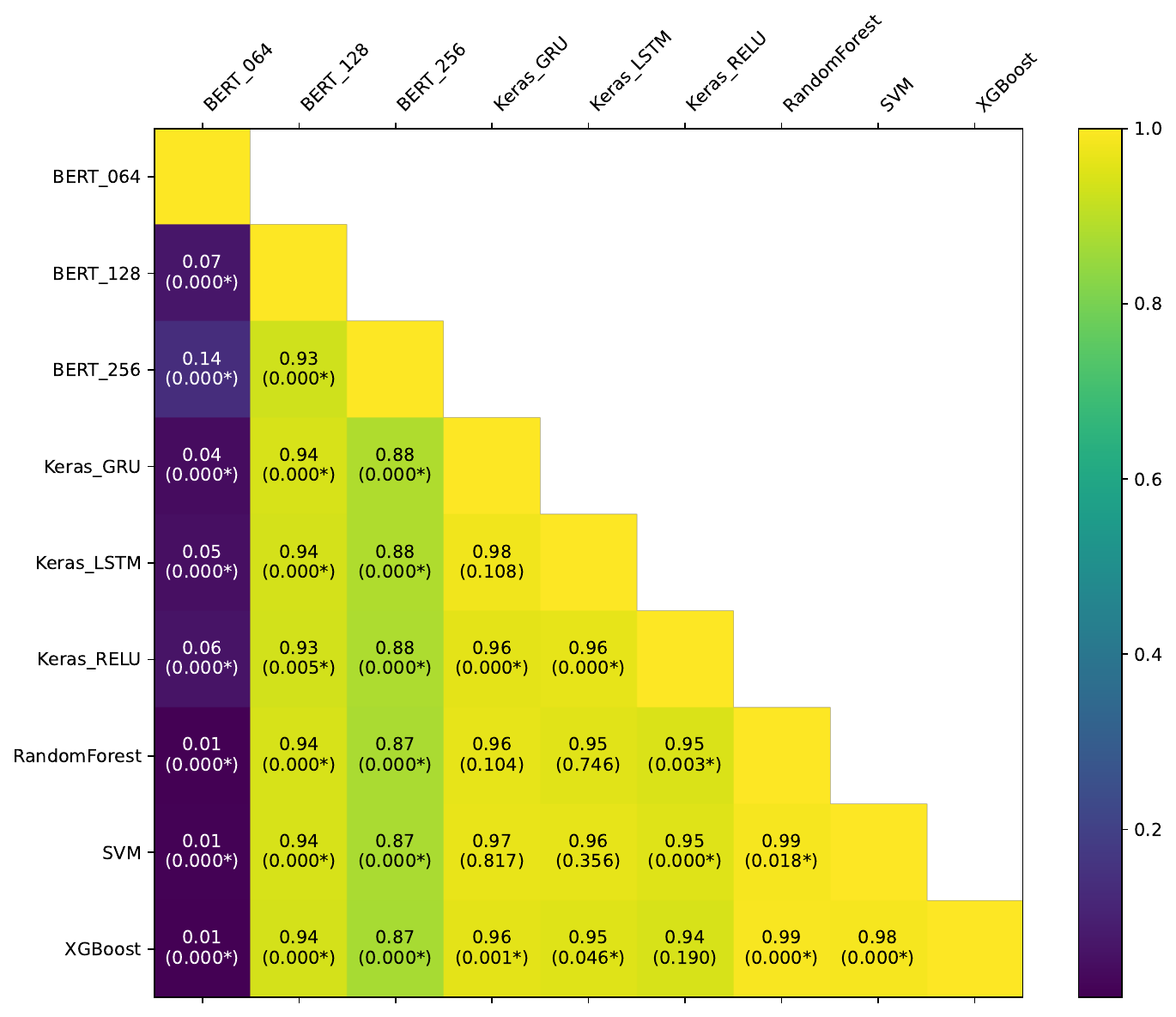}}
}
  \caption{Agreement matrices indicating the level of consistency between models with respect to their binary classification of narratives for (a) Birth event narratives, (b) Death event narratives, (c) Hired event narratives, and (d) Fired event narratives. *'s represent statistically significant results between the models.}
\label{fig:foobar}
\vskip -0.2in
\end{figure*}
\FloatBarrier 

\section{McNemar test results using the manually tagged data set.}
Figures 8-11 provide the results of the McNemar tests for significant differences in the distribution of binary Yes/No classifications of the narratives of each event type for each of the nine models. A significance level of 0.05 is assigned to assess significance. Within each figure, the red dotted line represents the significance level. Each plot within figures 8-11 displays the P-value resulting from a comparison of the model listed at the top of the plot against each of the other eight models. As such, each figure contains nine plots and each plot contains eight bars.

These comparisons are conducted using the confusion matrices calculated from each of the developed ML models' tagged data (n = 2,880) using 10 k-fold cross validation. In the confusion matrices, the true values are assigned based on the tagged data and the predicted values are based on the ML models' predicted tags. The null hypothesis for this set of tests is that there is no statistically significant difference exists between the distributions of the binary classifications made by each model (\textit{i.e.,} the distributions of Yes/No classifications is the same between ML models). The alternative hypothesis is that a statistically significant difference between the binary classifications does exist between the models (\textit{i.e.,} the distribution of Yes/No classifications is different between the ML models). P-values less than 0.05 indicate that evidence exists to support rejecting the null hypothesis in favor the alternative hypothesis that a difference between the distribution of binary classifications does exist.

\begin{figure}[ht]
\vskip 0.2in
\begin{center}
\centerline{\includegraphics[width=\columnwidth]{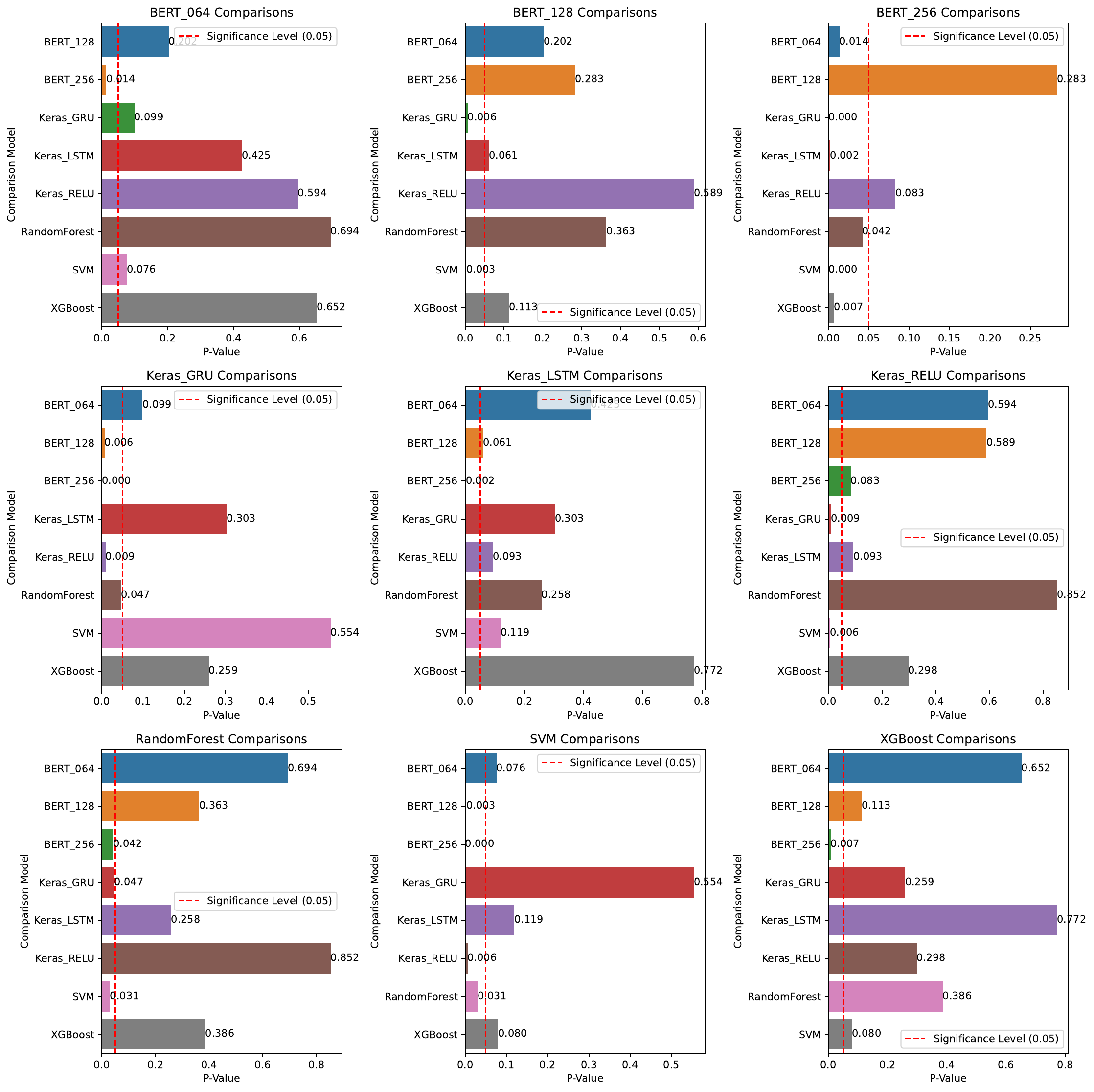}}
\caption{McNemar tests for significant differences in the distributions of \textit{Yes}/\textit{No} classifications of Birth narratives between each of the nine models. These comparisons utilize the manually tagged data set (n = 2,880) as the baseline for comparison.}
\label{icml-historical}
\end{center}
\vskip -0.2in
\end{figure}
\FloatBarrier 

\begin{figure}[ht]
\vskip 0.2in
\begin{center}
\centerline{\includegraphics[width=\columnwidth]{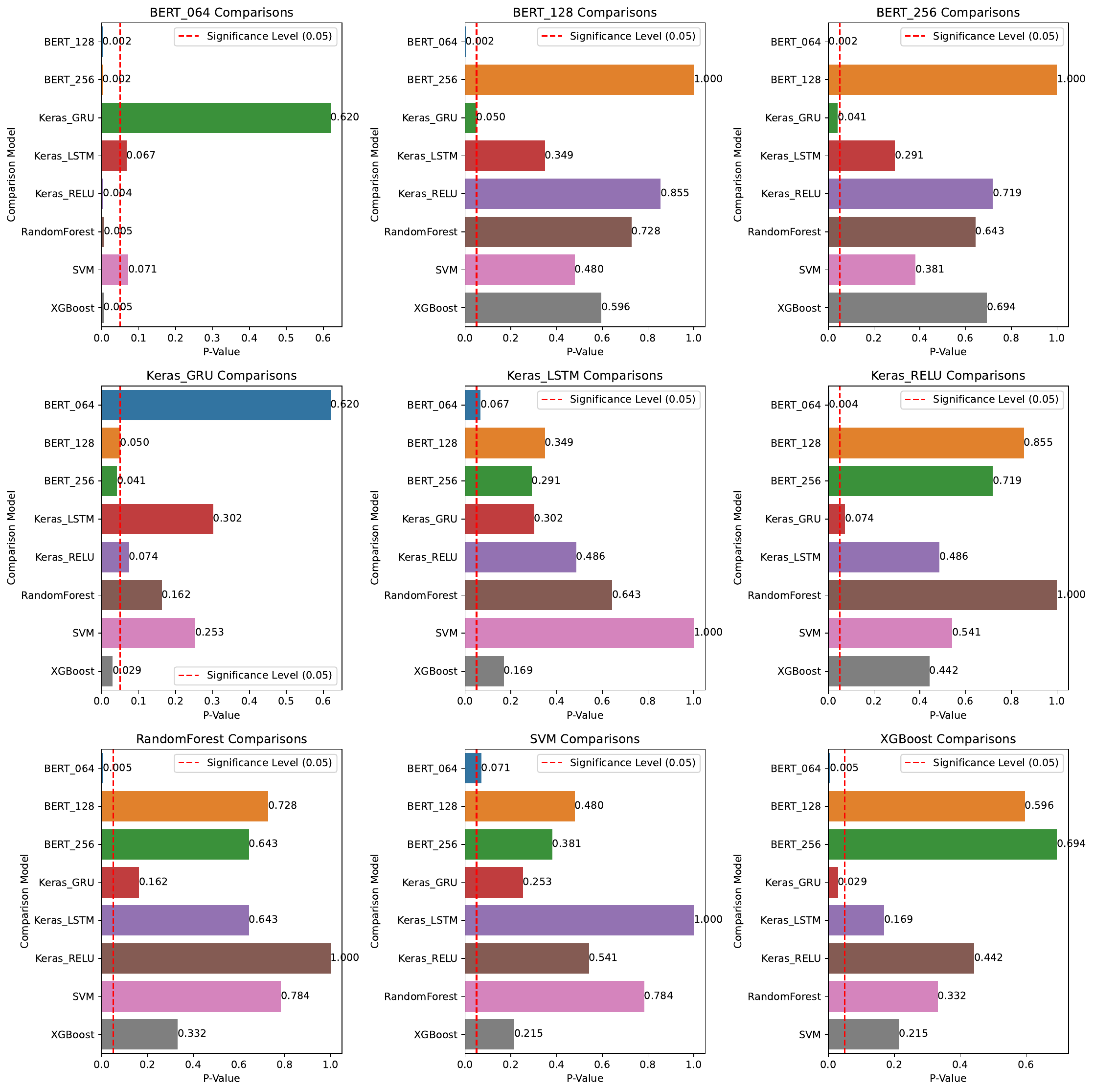}}
\caption{McNemar tests for significant differences in the distributions of \textit{Yes}/\textit{No} classifications of Death narratives between each of the nine models. These comparisons utilize the manually tagged data set (n = 2,880) as the baseline for comparison.}
\label{icml-historical}
\end{center}
\vskip -0.2in
\end{figure}
\FloatBarrier 

\begin{figure}[ht]
\vskip 0.2in
\begin{center}
\centerline{\includegraphics[width=\columnwidth]{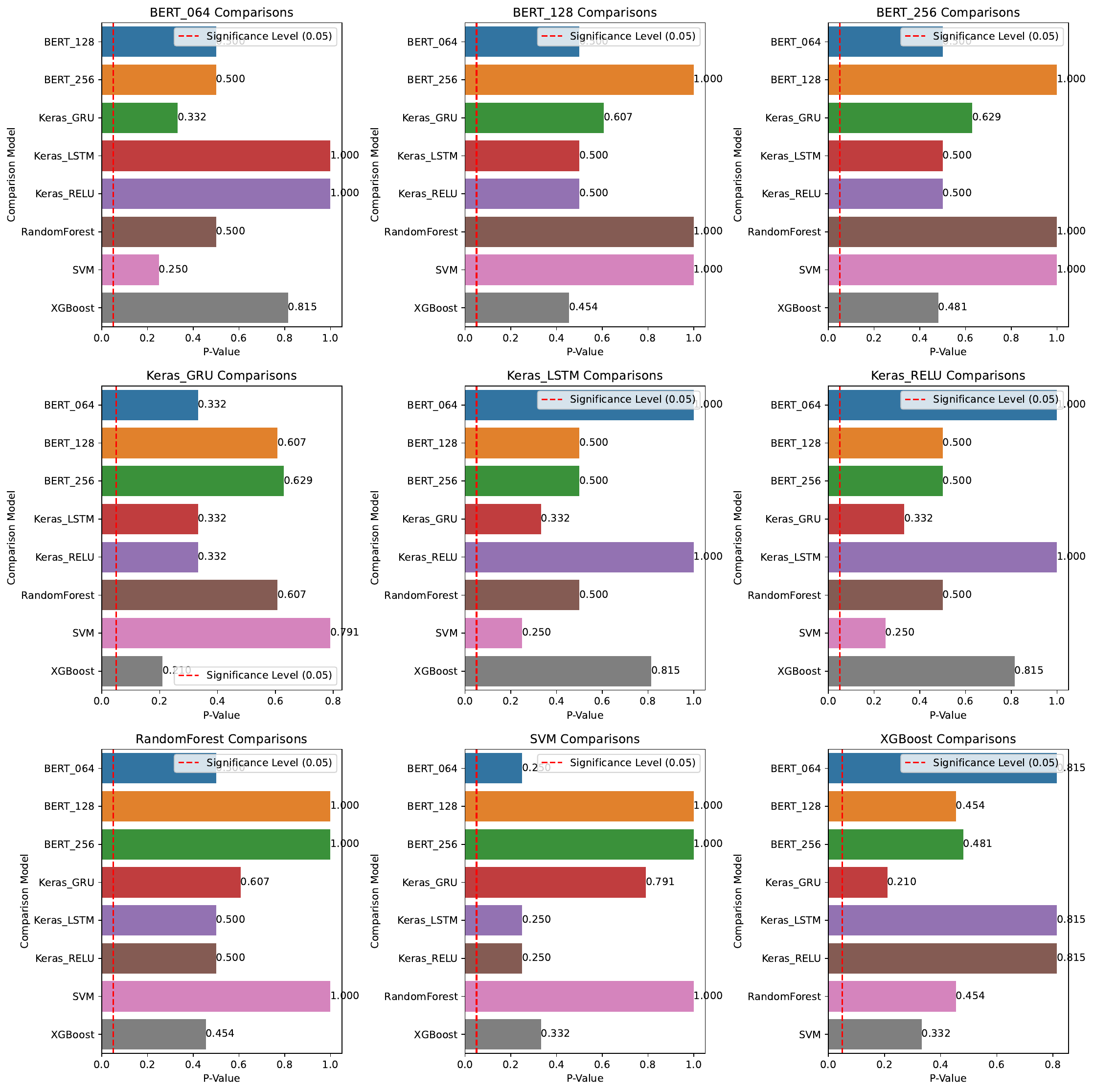}}
\caption{McNemar tests for significant differences in the distributions of \textit{Yes}/\textit{No} classifications of Hired narratives between each of the nine models. These comparisons utilize the manually tagged data set (n = 2,880) as the baseline for comparison.}
\label{icml-historical}
\end{center}
\vskip -0.2in
\end{figure}
\FloatBarrier 

\begin{figure}[ht]
\vskip 0.2in
\begin{center}
\centerline{\includegraphics[width=\columnwidth]{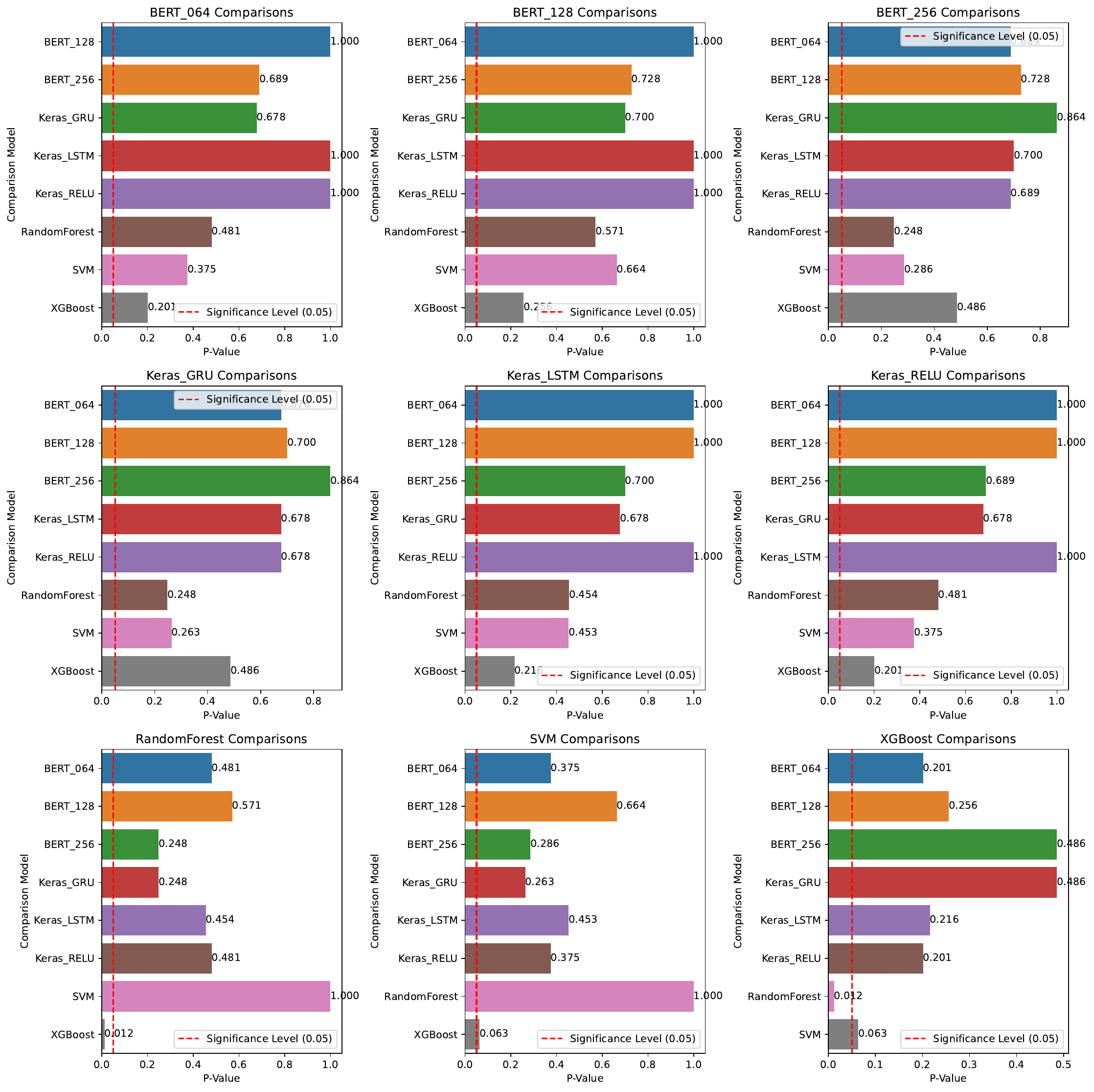}}
\caption{McNemar tests for significant differences in the distributions of \textit{Yes}/\textit{No} classifications of Fired narratives between each of the nine models. These comparisons utilize the manually tagged data set (n = 2,880) as the baseline for comparison.}
\label{icml-historical}
\end{center}
\vskip -0.2in
\end{figure}
\FloatBarrier 

\section{McNemar test results of predicted classifications on the originally untagged data set.}
Figures 12-15 provide the results of the McNemar tests for significant differences in the distribution of binary \textit{Yes}/\textit{No} narratives' predicted classifications of the originally untagged data set. A significance level of 0.05 is assigned to assess significance. Within each figure, the red dotted line represents the significance level. Each plot within each figure displays the P-value resulting from a comparison of the model listed at the top of the plot against each of the other eight models. This results in nine plots within each figure that each contain eight bars.

These comparisons are conducted using the originally untagged data (n = 21,120) using an ensemble hypothesis comparison set where baseline \textit{Yes} classifications are the result of 5+ models predicting a \textit{Yes}. The null hypothesis tested in these cases states that there is no statistically significant difference in distributions of the binary classifications made by each model. P-values less than 0.05 indicate that evidence exists to support rejecting the null hypothesis in favor the alternative hypothesis that a difference between the distribution of binary classifications does exist.

\begin{figure}[ht]
\vskip 0.2in
\begin{center}
\centerline{\includegraphics[width=\columnwidth]{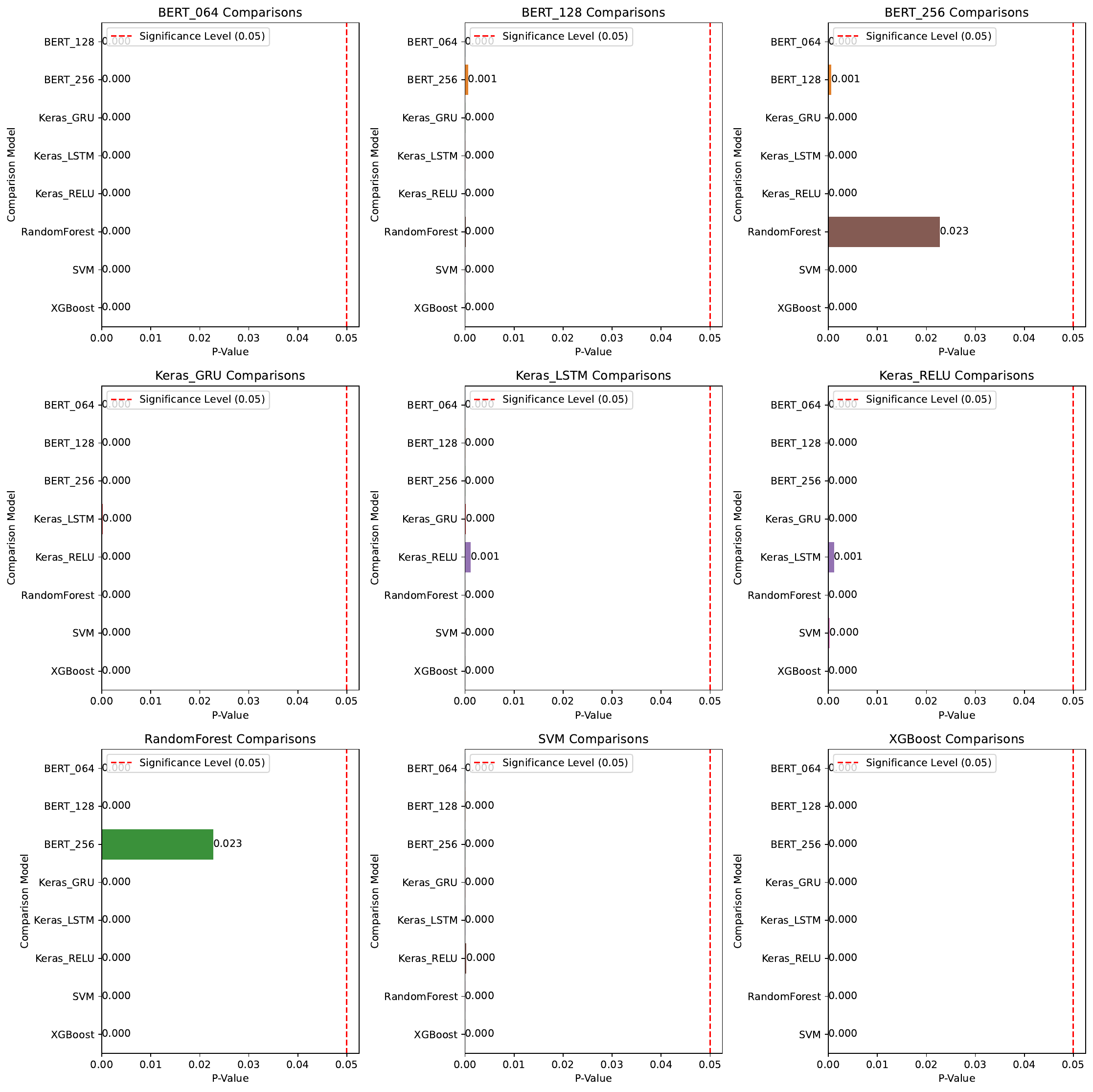}}
\caption{McNemar tests for significant differences in the distributions of \textit{Yes}/\textit{No} predicted classifications of Birth narratives between each of the nine models. These comparisons utilize the ensemble hypothesis data set (n = 21,120) as the baseline for comparison.}
\label{icml-historical}
\end{center}
\vskip -0.2in
\end{figure}
\FloatBarrier 

\begin{figure}[ht]
\vskip 0.2in
\begin{center}
\centerline{\includegraphics[width=\columnwidth]{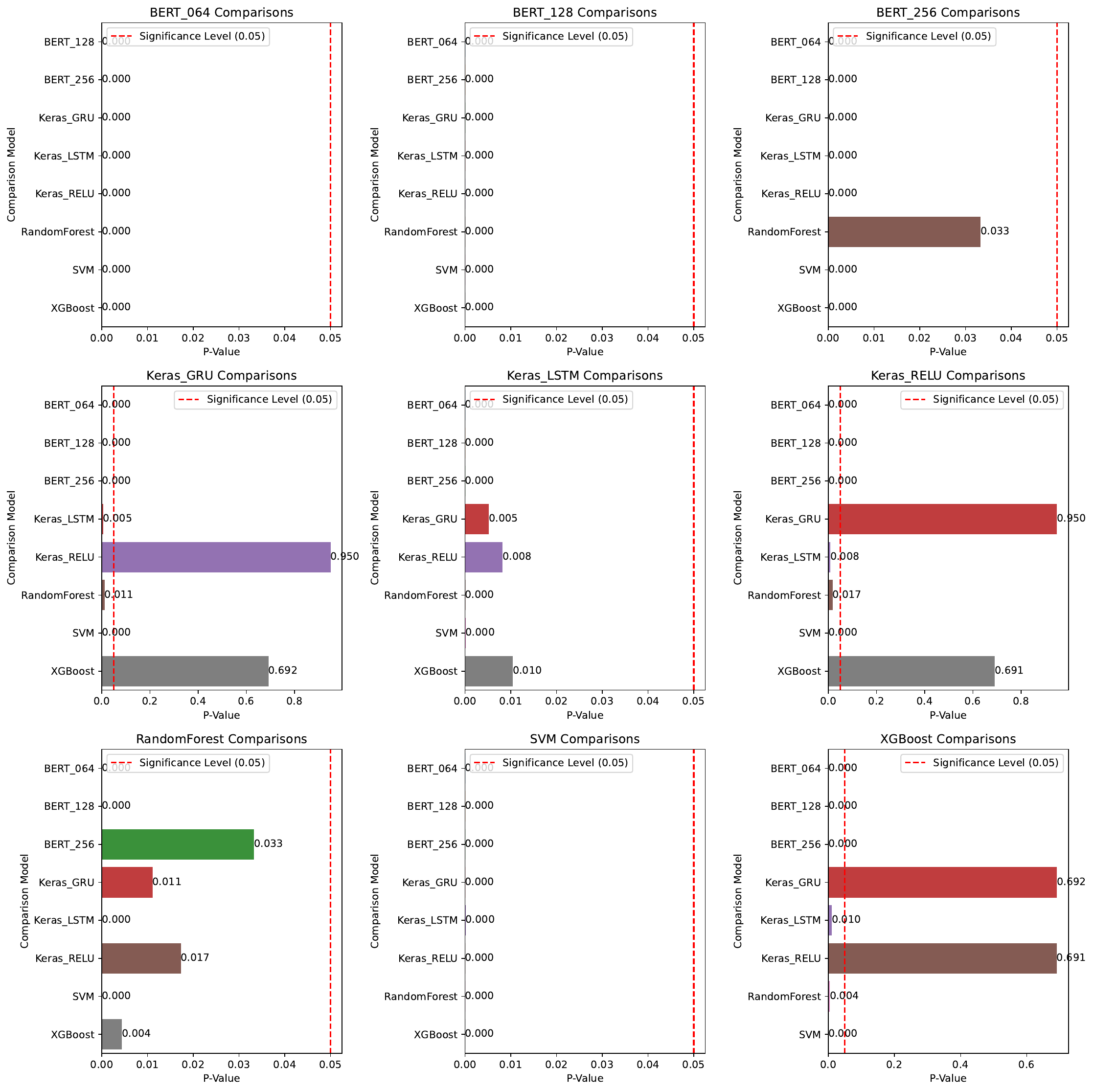}}
\caption{McNemar tests for significant differences in the distributions of \textit{Yes}/\textit{No} predicted classifications of Death narratives between each of the nine models. These comparisons utilize the ensemble hypothesis data set (n = 21,120) as the baseline for comparison.}
\label{icml-historical}
\end{center}
\vskip -0.2in
\end{figure}
\FloatBarrier 

\begin{figure}[ht]
\vskip 0.2in
\begin{center}
\centerline{\includegraphics[width=\columnwidth]{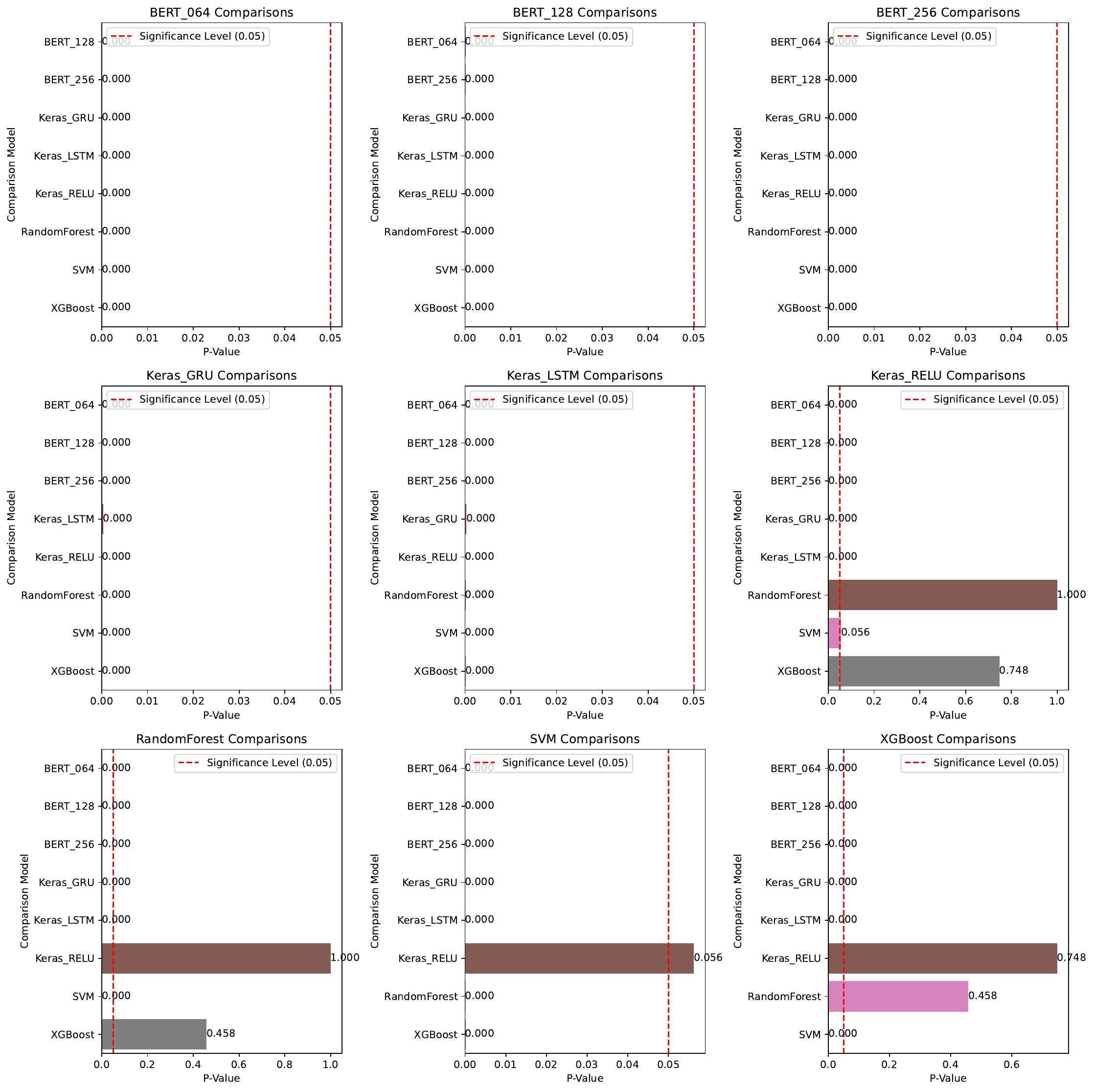}}
\caption{McNemar tests for significant differences in the distributions of \textit{Yes}/\textit{No} predicted classifications of Hired narratives between each of the nine models. These comparisons utilize the ensemble hypothesis data set (n = 21,120) as the baseline for comparison.}
\label{icml-historical}
\end{center}
\vskip -0.2in
\end{figure}
\FloatBarrier 

\begin{figure}[ht]
\vskip 0.2in
\begin{center}
\centerline{\includegraphics[width=\columnwidth]{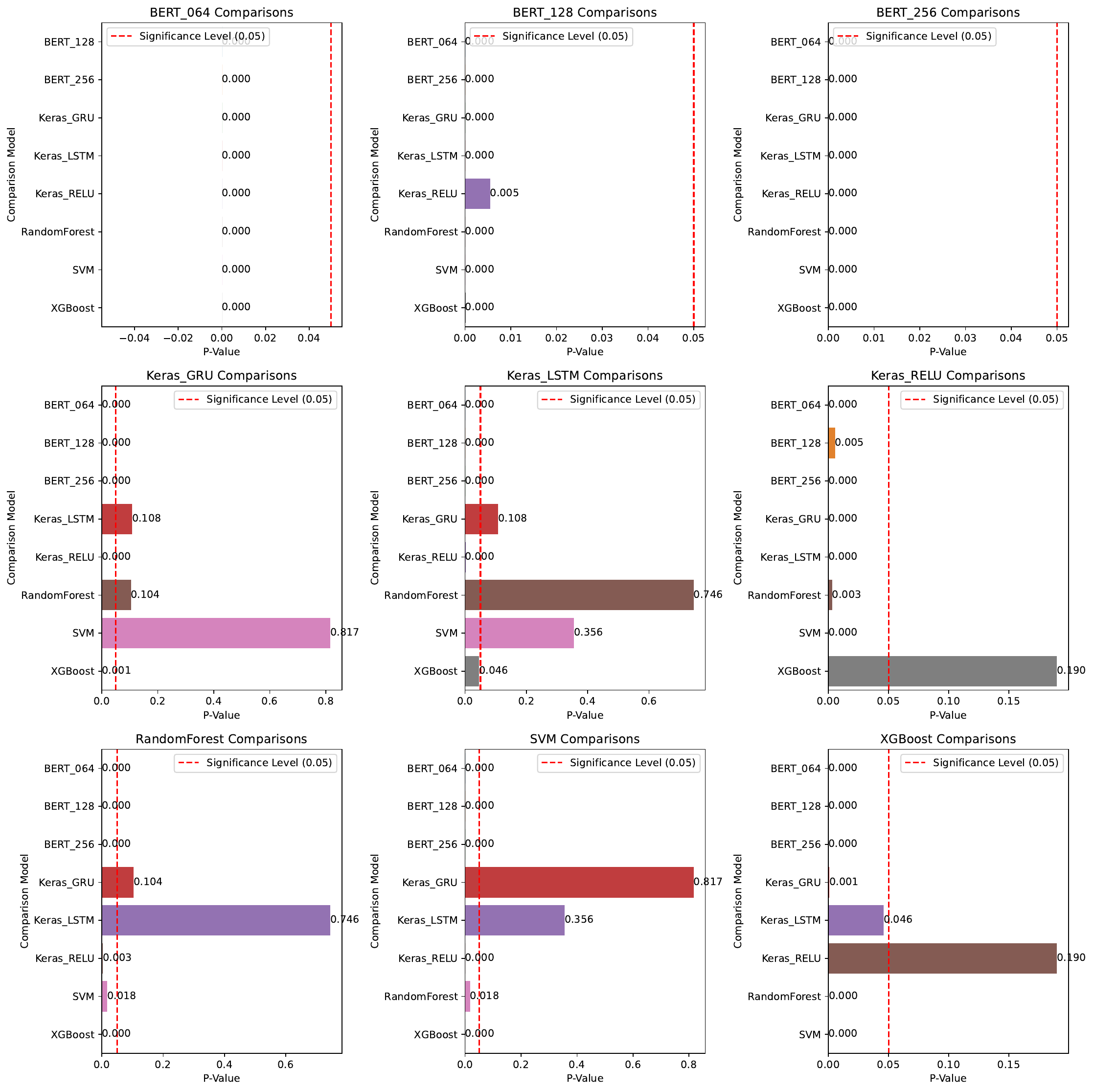}}
\caption{McNemar tests for significant differences in the distributions of \textit{Yes}/\textit{No} predicted classifications of Fired narratives between each of the nine models. These comparisons utilize the ensemble hypothesis data set (n = 21,120) as the baseline for comparison.}
\label{icml-historical}
\end{center}
\vskip -0.2in
\end{figure}
\FloatBarrier 

\section{Confusion matrices for ML models.}
Figures 16-24 provide the confusion matrices for each of the developed ML models. Each figure is divided into four parts: (a) confusion matrix for models built using Birth event narratives, (b) confusion matrix for models built using Death event narratives, (c) confusion matrix for models built using Hired event narratives, and (d) confusion matrix for models built using Fired event narratives. These confusion matrices convey the number of True Positives, False Positives, True Negatives, and False Negatives for each of the developed models based on how well the models matched the 2,880 tagged data points. Each figure provides the sample size and normalized percentage of samples within each cell of the confusion matrix.

\begin{figure*}[ht]
\vskip 0.2in
\centerline{
  \subfigure(a){\includegraphics[width=0.5\textwidth, height=150pt]{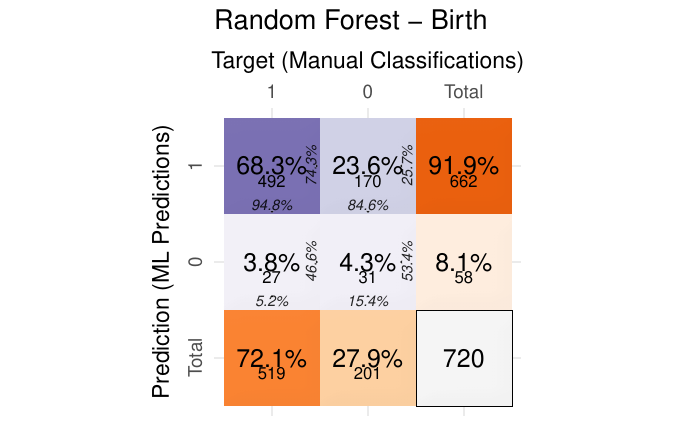}} 
  \subfigure(b){\includegraphics[width=0.5\textwidth, height=150pt]{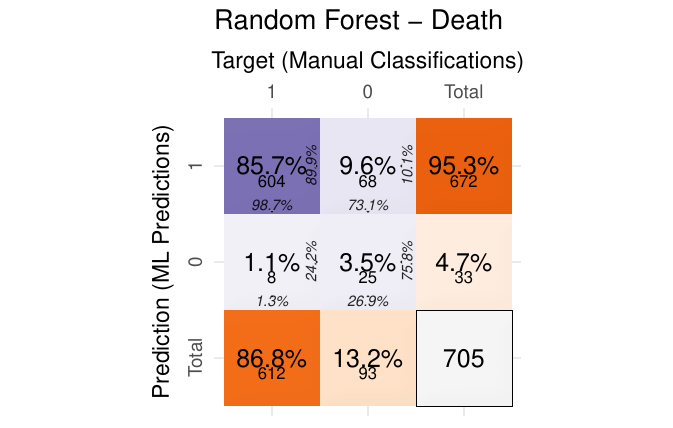}} 
  }
\centerline{
\subfigure(c){\includegraphics[width=0.5\textwidth, height=150pt]{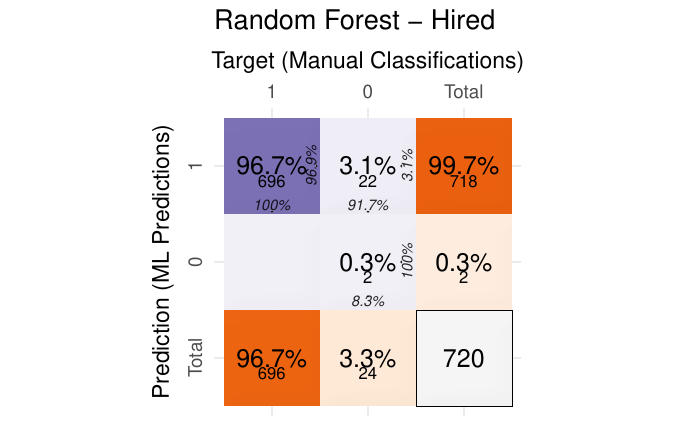}}
  \subfigure(d){\includegraphics[width=0.5\textwidth, height=150pt]{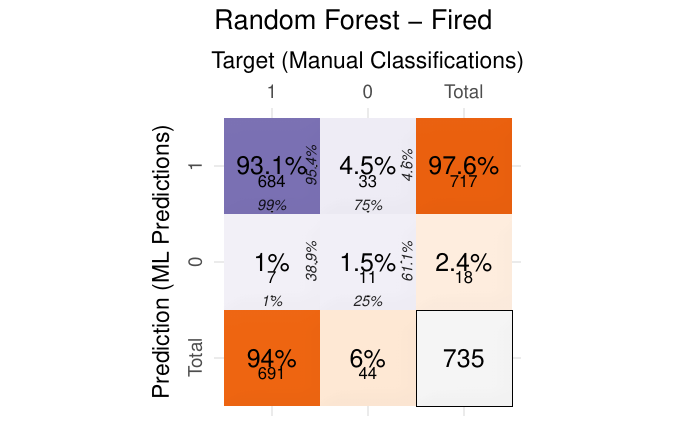}}
}
  \caption{Confusion matrices for Random Forest models indicating the number of True Positives, False Positives, True Negatives, and False Negatives with respect to their binary classification of narratives for (a) Birth event narratives, (b) Death event narratives, (c) Hired event narratives, and (d) Fired event narratives.}
\label{fig:cm_rf}
\vskip -0.2in
\end{figure*}
\FloatBarrier 


\begin{figure*}[ht]
\vskip 0.2in
\centerline{
  \subfigure(a){\includegraphics[width=0.5\textwidth, height=150pt]{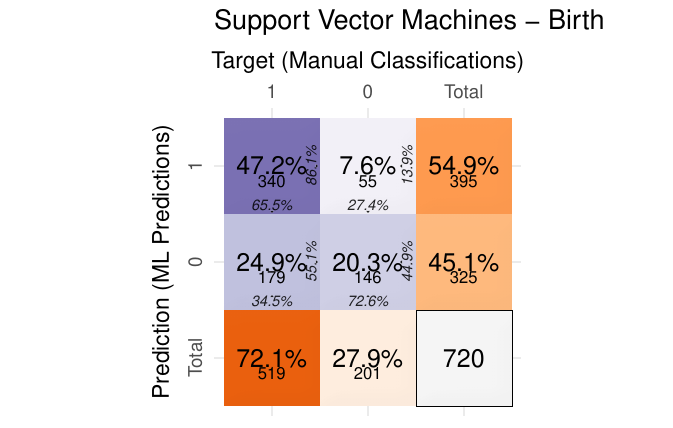}} 
  \subfigure(b){\includegraphics[width=0.5\textwidth, height=150pt]{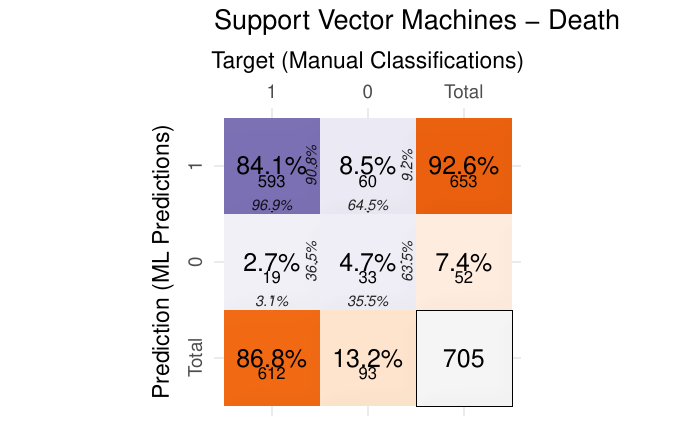}} 
  }
\centerline{
\subfigure(c){\includegraphics[width=0.5\textwidth, height=150pt]{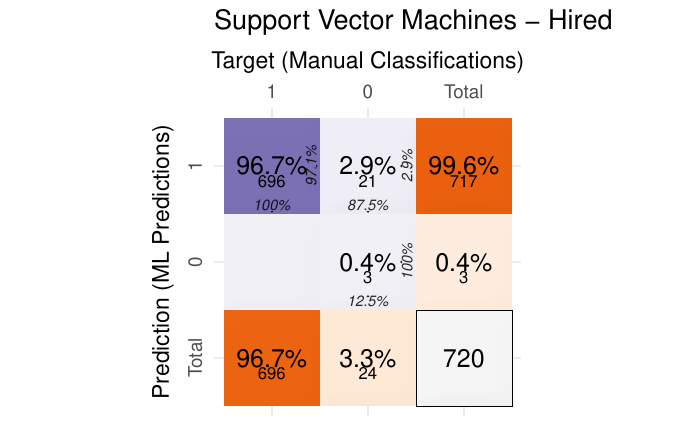}}
  \subfigure(d){\includegraphics[width=0.5\textwidth, height=150pt]{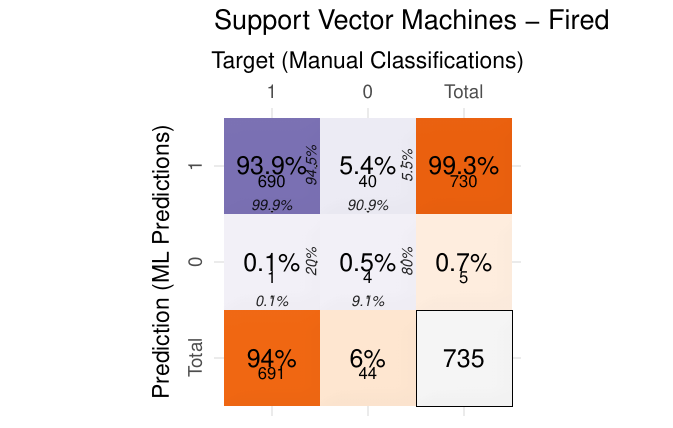}}
}
  \caption{Confusion matrices for Support Vector Machine (SVM) models indicating the number of True Positives, False Positives, True Negatives, and False Negatives with respect to their binary classification of narratives for (a) Birth event narratives, (b) Death event narratives, (c) Hired event narratives, and (d) Fired event narratives.}
\label{fig:cm_svm}
\vskip -0.2in
\end{figure*}
\FloatBarrier 


\begin{figure*}[ht]
\vskip 0.2in
\centerline{
  \subfigure(a){\includegraphics[width=0.5\textwidth, height=150pt]{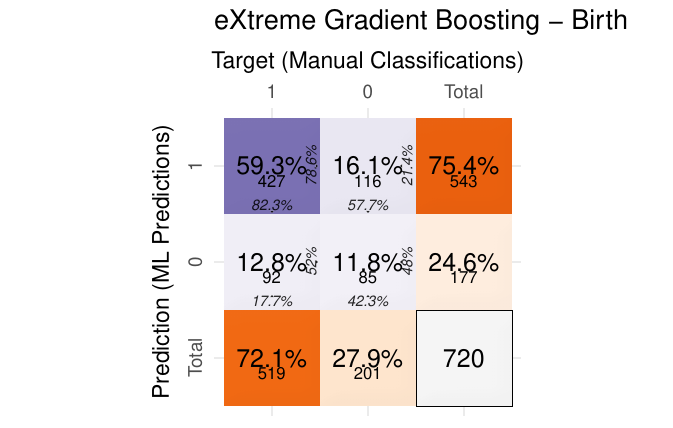}} 
  \subfigure(b){\includegraphics[width=0.5\textwidth, height=150pt]{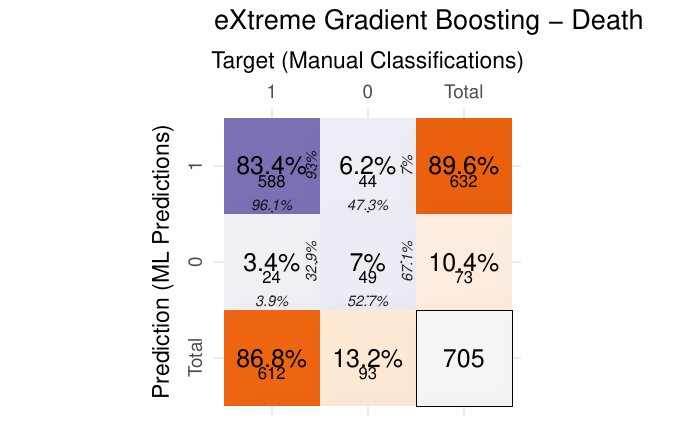}} 
  }
\centerline{
\subfigure(c){\includegraphics[width=0.5\textwidth, height=150pt]{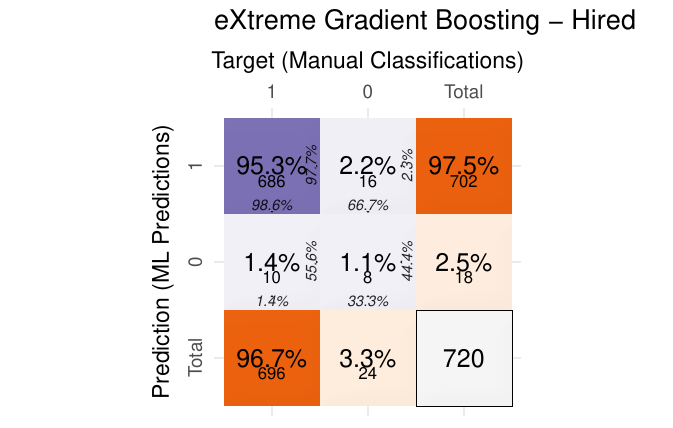}}
  \subfigure(d){\includegraphics[width=0.5\textwidth, height=150pt]{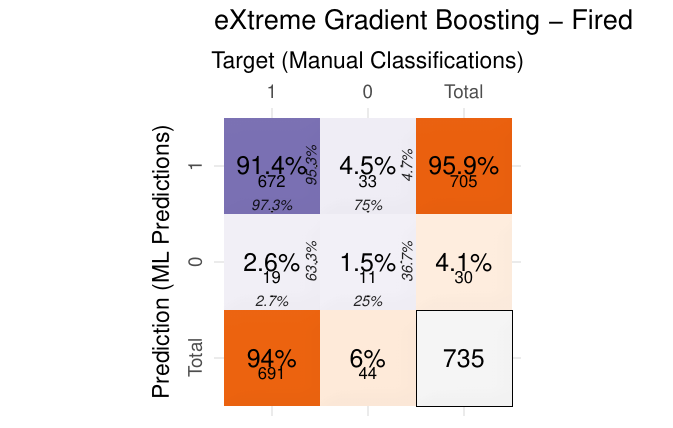}}
}
  \caption{Confusion matrices for eXtreme Gradient Boosting models indicating the number of True Positives, False Positives, True Negatives, and False Negatives with respect to their binary classification of narratives for (a) Birth event narratives, (b) Death event narratives, (c) Hired event narratives, and (d) Fired event narratives.}
\label{fig:cm_xgboost}
\vskip -0.2in
\end{figure*}
\FloatBarrier 


\begin{figure*}[ht]
\vskip 0.2in
\centerline{
  \subfigure(a){\includegraphics[width=0.5\textwidth, height=150pt]{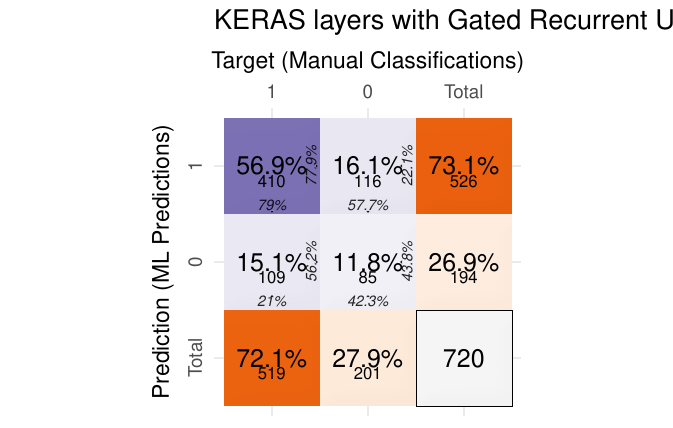}} 
  \subfigure(b){\includegraphics[width=0.5\textwidth, height=150pt]{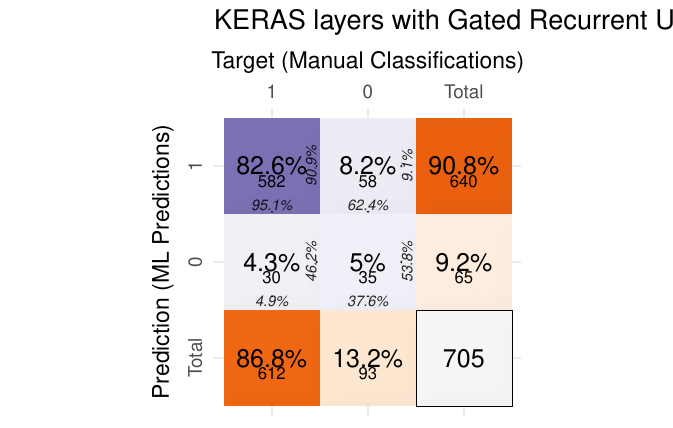}} 
  }
\centerline{
\subfigure(c){\includegraphics[width=0.5\textwidth, height=150pt]{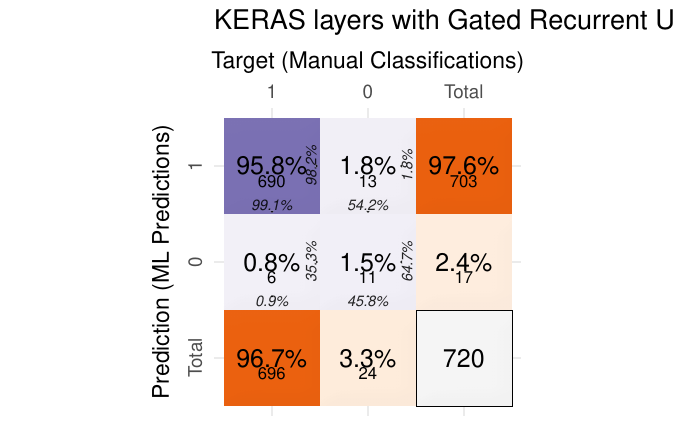}}
  \subfigure(d){\includegraphics[width=0.5\textwidth, height=150pt]{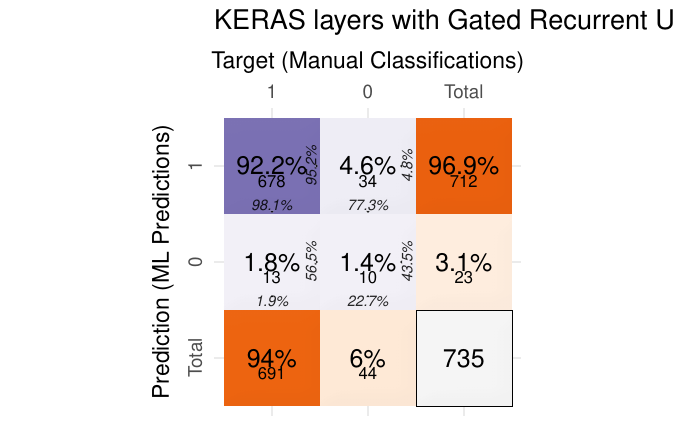}}
}
  \caption{Confusion matrices for KERAS layers with Gated Recurrent Unit (GRU) models indicating the number of True Positives, False Positives, True Negatives, and False Negatives with respect to their binary classification of narratives for (a) Birth event narratives, (b) Death event narratives, (c) Hired event narratives, and (d) Fired event narratives.}
\label{fig:cm_kgru}
\vskip -0.2in
\end{figure*}
\FloatBarrier 

\begin{figure*}[ht]
\vskip 0.2in
\centerline{
  \subfigure(a){\includegraphics[width=0.5\textwidth, height=150pt]{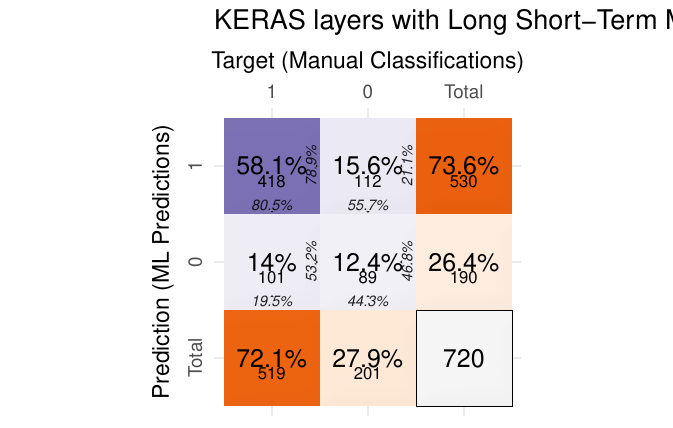}} 
  \subfigure(b){\includegraphics[width=0.5\textwidth, height=150pt]{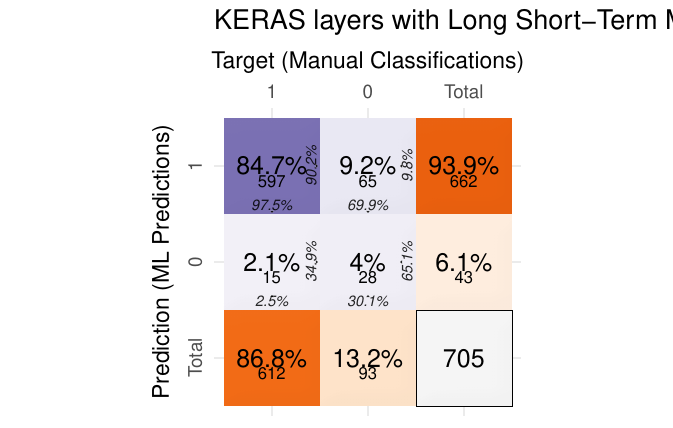}} 
  }
\centerline{
\subfigure(c){\includegraphics[width=0.5\textwidth, height=150pt]{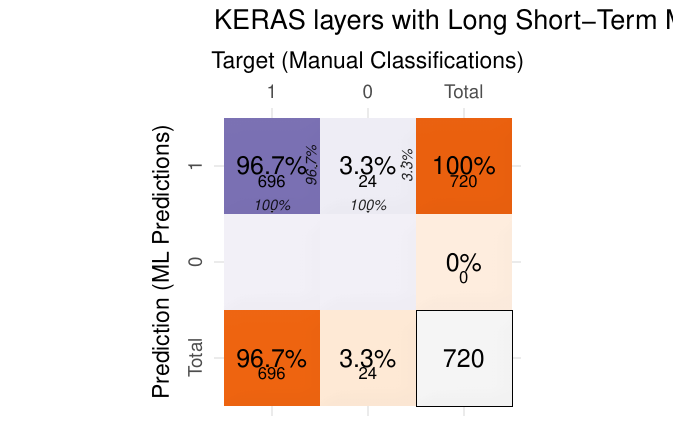}}
  \subfigure(d){\includegraphics[width=0.5\textwidth, height=150pt]{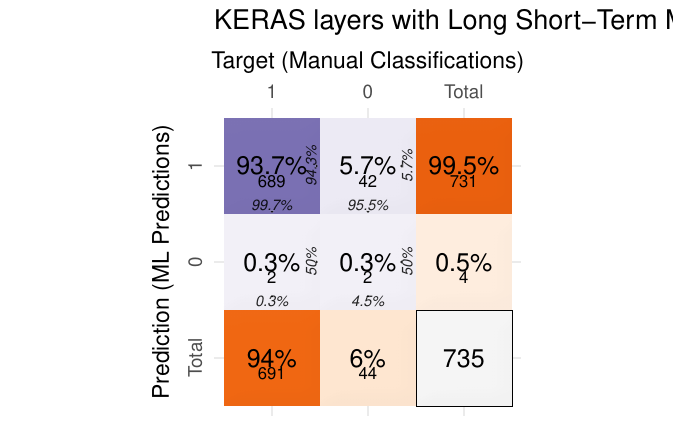}}
}
  \caption{Confusion matrices for KERAS layers with Long Short-Term Memory (LSTM) models indicating the number of True Positives, False Positives, True Negatives, and False Negatives with respect to their binary classification of narratives for (a) Birth event narratives, (b) Death event narratives, (c) Hired event narratives, and (d) Fired event narratives.}
\label{fig:cm_klstm}
\vskip -0.2in
\end{figure*}
\FloatBarrier 

\begin{figure*}[ht]
\vskip 0.2in
\centerline{
  \subfigure(a){\includegraphics[width=0.5\textwidth, height=150pt]{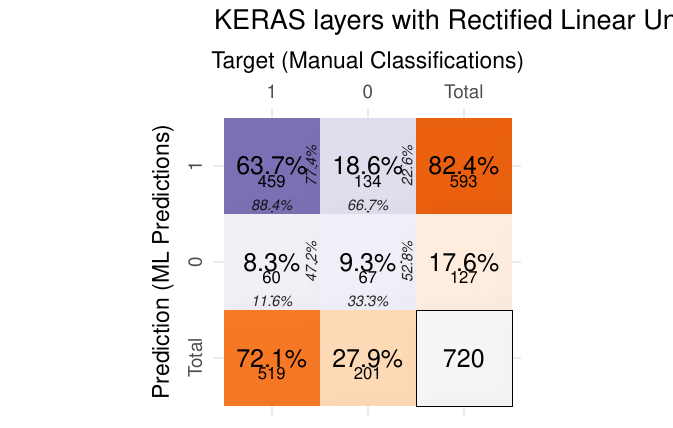}} 
  \subfigure(b){\includegraphics[width=0.5\textwidth, height=150pt]{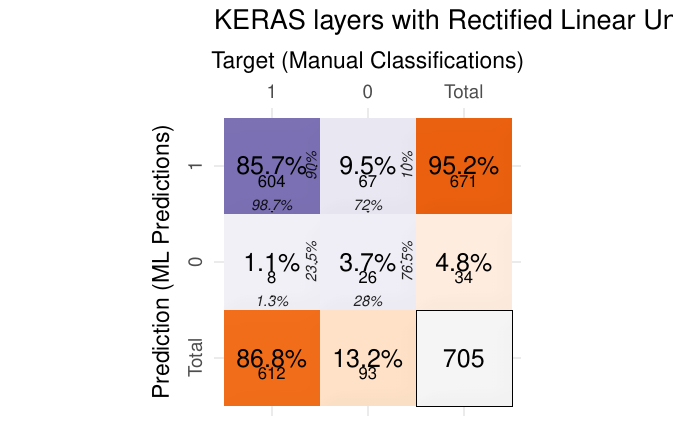}} 
  }
\centerline{
\subfigure(c){\includegraphics[width=0.5\textwidth, height=150pt]{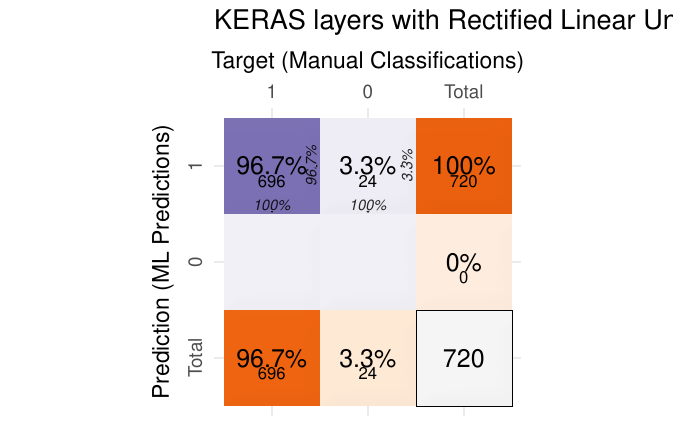}}
  \subfigure(d){\includegraphics[width=0.5\textwidth, height=150pt]{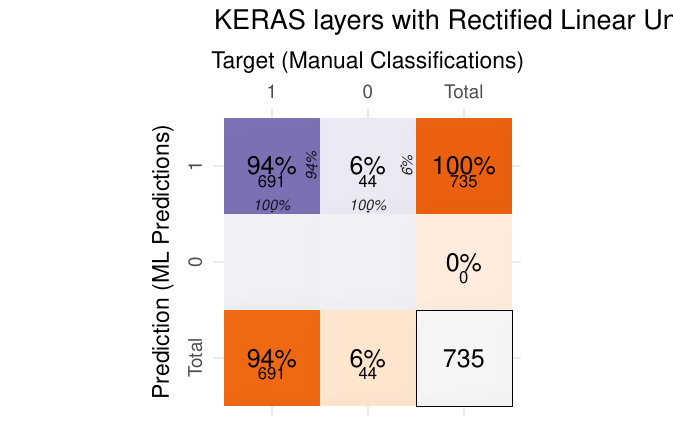}}
}
  \caption{Confusion matrices for KERAS layers with Rectified Linear Unit (RELU) models indicating the number of True Positives, False Positives, True Negatives, and False Negatives with respect to their binary classification of narratives for (a) Birth event narratives, (b) Death event narratives, (c) Hired event narratives, and (d) Fired event narratives.}
\label{fig:cm_krelu}
\vskip -0.2in
\end{figure*}
\FloatBarrier 

\begin{figure*}[ht]
\vskip 0.2in
\centerline{
  \subfigure(a){\includegraphics[width=0.5\textwidth, height=150pt]{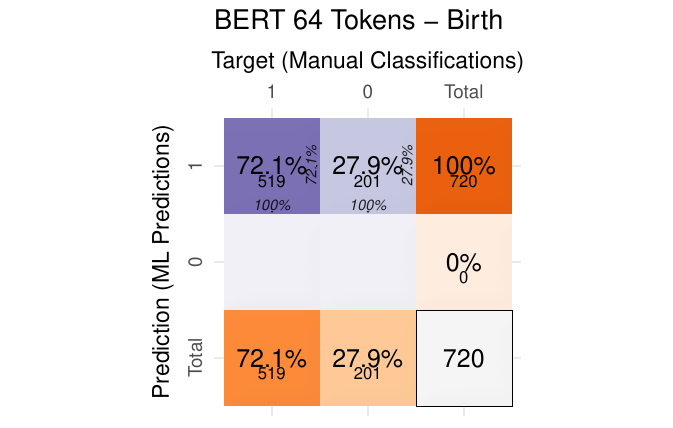}} 
  \subfigure(b){\includegraphics[width=0.5\textwidth, height=150pt]{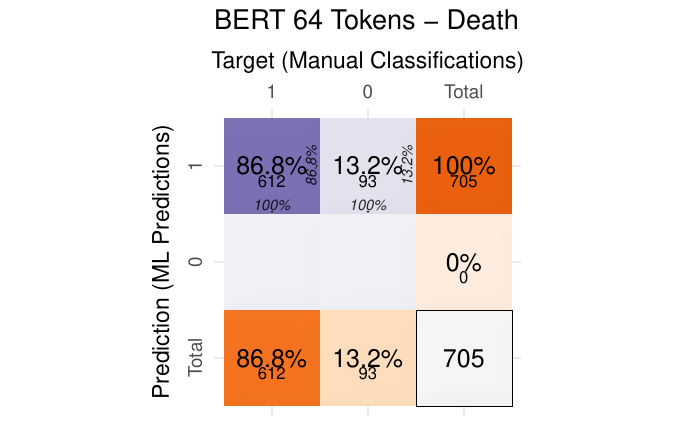}} 
  }
\centerline{
\subfigure(c){\includegraphics[width=0.5\textwidth, height=150pt]{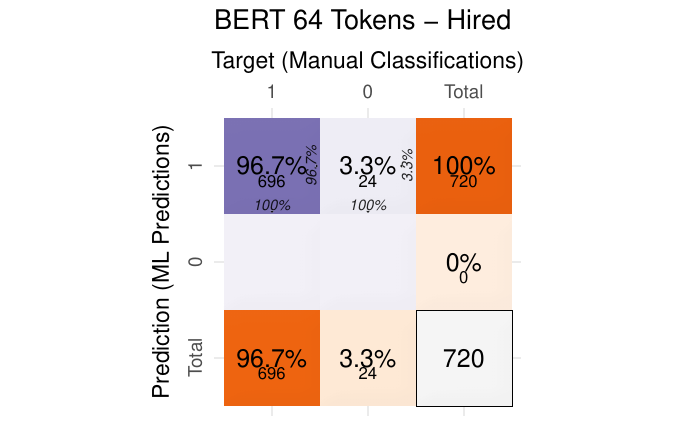}}
  \subfigure(d){\includegraphics[width=0.5\textwidth, height=150pt]{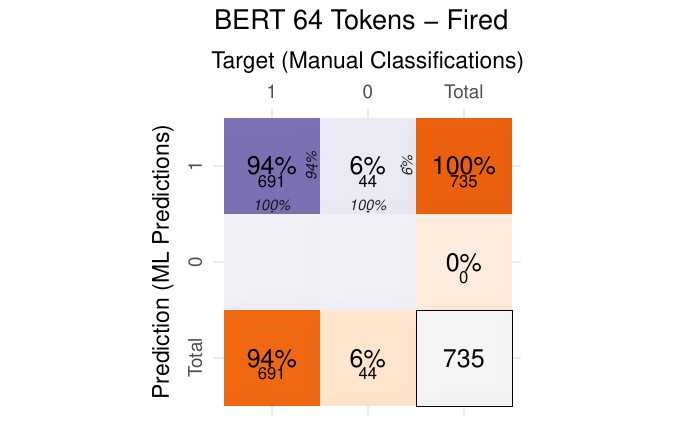}}
}
  \caption{Confusion matrices for Bidirectional Encoder Representations from Transformers (BERT) with 64 token limit models. The matrices indicate the number of True Positives, False Positives, True Negatives, and False Negatives with respect to their binary classification of narratives for (a) Birth event narratives, (b) Death event narratives, (c) Hired event narratives, and (d) Fired event narratives.}
\label{fig:cm_bert64}
\vskip -0.2in
\end{figure*}
\FloatBarrier 


\begin{figure*}[ht]
\vskip 0.2in
\centerline{
  \subfigure(a){\includegraphics[width=0.5\textwidth, height=150pt]{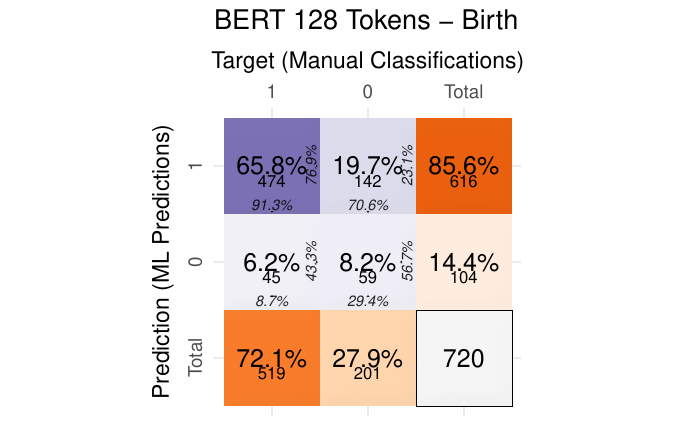}} 
  \subfigure(b){\includegraphics[width=0.5\textwidth, height=150pt]{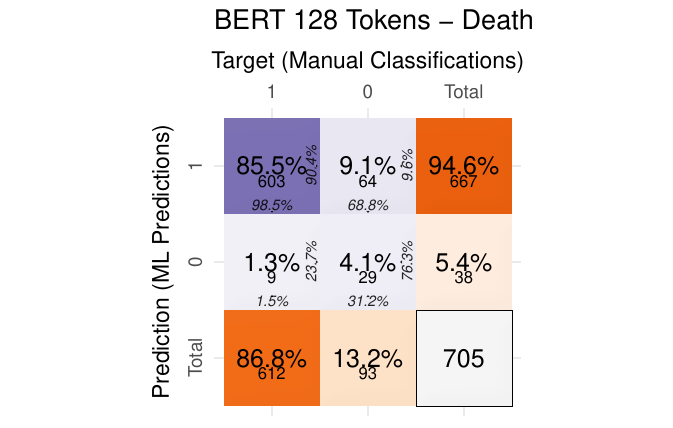}} 
  }
\centerline{
\subfigure(c){\includegraphics[width=0.5\textwidth, height=150pt]{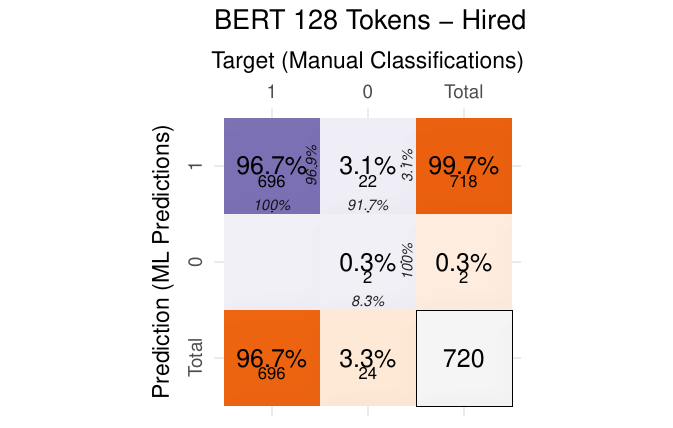}}
  \subfigure(d){\includegraphics[width=0.5\textwidth, height=150pt]{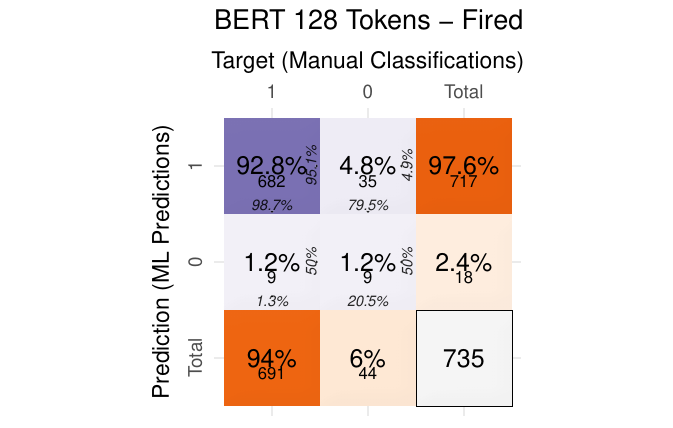}}
}
  \caption{Confusion matrices for Bidirectional Encoder Representations from Transformers (BERT) with 128 token limit models. The matrices indicate the number of True Positives, False Positives, True Negatives, and False Negatives with respect to their binary classification of narratives for (a) Birth event narratives, (b) Death event narratives, (c) Hired event narratives, and (d) Fired event narratives.}
\label{fig:cm_bert128}
\vskip -0.2in
\end{figure*}
\FloatBarrier 


\begin{figure*}[ht]
\vskip 0.2in
\centerline{
  \subfigure(a){\includegraphics[width=0.5\textwidth, height=150pt]{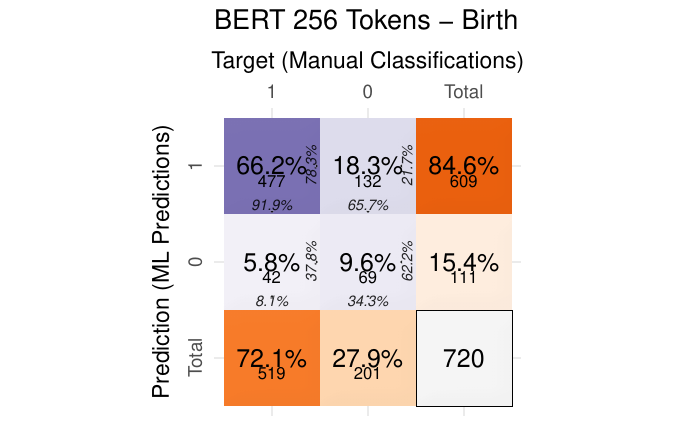}} 
  \subfigure(b){\includegraphics[width=0.5\textwidth, height=150pt]{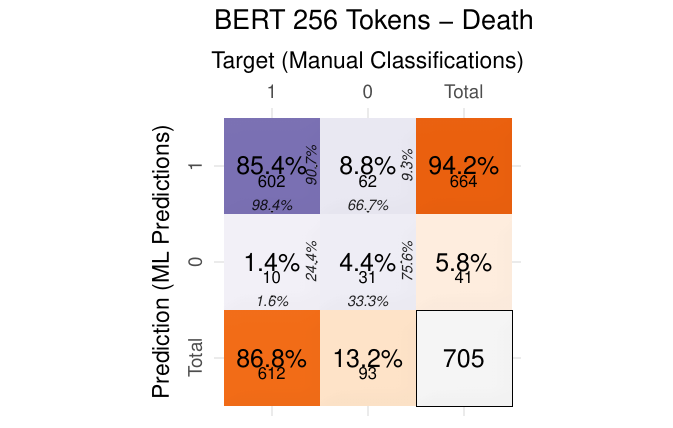}} 
  }
\centerline{
\subfigure(c){\includegraphics[width=0.5\textwidth, height=150pt]{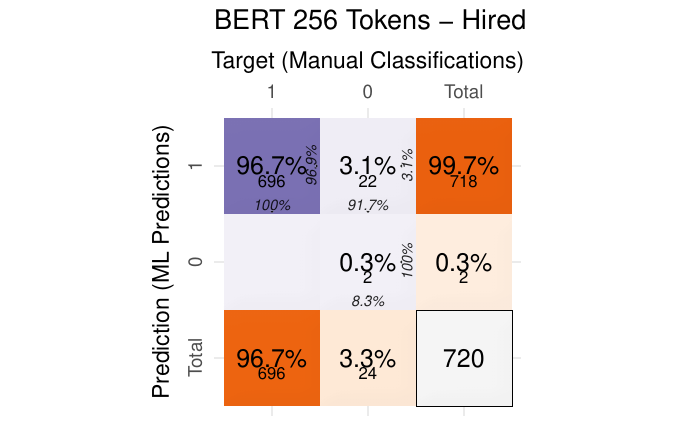}}
  \subfigure(d){\includegraphics[width=0.5\textwidth, height=150pt]{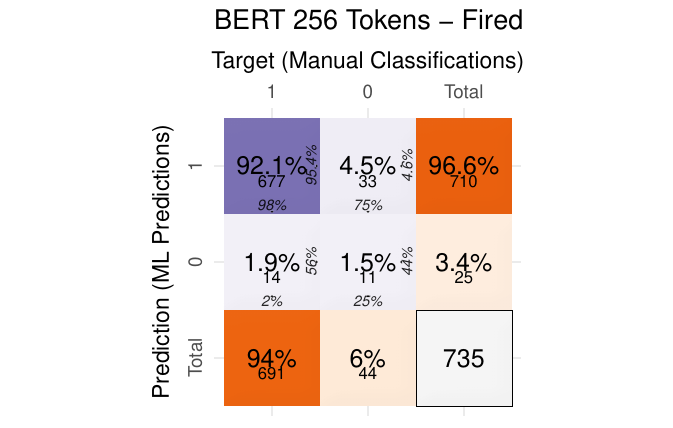}}
}
  \caption{Confusion matrices for Bidirectional Encoder Representations from Transformers (BERT) with 256 token limit models. The matrices indicate the number of True Positives, False Positives, True Negatives, and False Negatives with respect to their binary classification of narratives for (a) Birth event narratives, (b) Death event narratives, (c) Hired event narratives, and (d) Fired event narratives.}
\label{fig:cm_bert256}
\vskip -0.2in
\end{figure*}
\FloatBarrier 

\end{document}